\documentclass{article}

\usepackage{microtype}
\usepackage{graphicx}
\usepackage{subcaption}
\usepackage{booktabs} 

\usepackage{hyperref}

\usepackage[preprint]{icml2026}

\usepackage{url}
\usepackage{booktabs}
\usepackage{algorithm}
\usepackage{algorithmic}
\usepackage{amsmath}
\usepackage{tikz}
\usepackage{xcolor}
\usepackage{subcaption}
\usepackage{siunitx}
\usepackage{threeparttable}
\usepackage{adjustbox}    
\usepackage{rotating}
\usepackage{caption}
\usepackage{xcolor}
\usepackage{float}
\usepackage{booktabs}
\usepackage{threeparttable}
\usepackage{graphicx}   
\usepackage{caption}
\usepackage{siunitx}
\usepackage{enumitem}
\usepackage{wrapfig} 
\usepackage{graphicx}

\usepackage{hyperref}
\usepackage{amsfonts}

\usepackage[capitalize,noabbrev]{cleveref}

\newtheorem{theorem}{Theorem}[section]
\newtheorem{proposition}[theorem]{Proposition}

\newtheorem{definition}[theorem]{Definition}

\usepackage[textsize=tiny]{todonotes}

\icmltitlerunning{Calibrated Multivariate Distributional Regression with Pre-Rank Regularization}

\begin{document}

\twocolumn[
  \icmltitle{Calibrated Multivariate Distributional Regression with Pre-Rank Regularization}


  \icmlsetsymbol{equal}{*}

  \begin{icmlauthorlist}
    \icmlauthor{Aya Laajil}{mbzuai}
    \icmlauthor{Elnura Zhalieva}{mbzuai}
    \icmlauthor{Naomi Desobry}{mbzuai}
    \icmlauthor{Souhaib Ben Taieb}{mbzuai}
  \end{icmlauthorlist}

  \icmlaffiliation{mbzuai}{Department of Statistics and Data Science, Mohamed Bin Zayed University of Artificial Intelligence, Abu Dhabi, UAE}

  \icmlcorrespondingauthor{Aya Laajil}{aya.laajil@mbzuai.ac.ae}
  \icmlcorrespondingauthor{Souhaib Ben Taieb}{souhaib.bentaieb@mbzuai.ac.ae}

  \icmlkeywords{Machine Learning, ICML}

  \vskip 0.3in
]




\printAffiliationsAndNotice{}


\begin{abstract}
The goal of probabilistic prediction is to issue predictive distributions that are as informative as possible, subject to being \textit{calibrated}. Despite substantial progress in the univariate setting, achieving multivariate calibration remains challenging. Recent work has introduced pre-rank functions, scalar projections of multivariate forecasts and observations, as flexible diagnostics for assessing specific aspects of multivariate calibration, but their use has largely been limited to post-hoc evaluation.
We propose a regularization-based calibration method that enforces multivariate calibration during training of multivariate distributional regression models using pre-rank functions. We further introduce a novel PCA-based pre-rank that projects predictions onto principal directions of the predictive distribution.
Through simulation studies and experiments on 18 real-world multi-output regression datasets, we show that the proposed approach substantially improves multivariate pre-rank calibration without compromising predictive accuracy, and that the PCA pre-rank reveals dependence-structure misspecifications that are not detected by existing pre-ranks.

\end{abstract}

\section{Introduction}

Probabilistic models aim to characterize uncertainty by issuing full predictive distributions rather than point estimates. The importance of \textit{calibration} has been demonstrated across a wide range of applications \citep{gneiting2014calibration, gulshan2016development, malik2019calibrated}. 
Because predictive distributions are often used directly for risk quantification and downstream decision making, statistical consistency between predicted distributions and observed outcomes is a fundamental requirement.
The goal of probabilistic modeling is therefore to learn predictive distributions that are as sharp as possible while being calibrated \citep{Gneiting2007}. This becomes particularly challenging in multivariate settings, where the joint dependence structure must also be captured. 

While probabilistic models are typically trained by minimizing a proper scoring rule, they can still be miscalibrated particularly under limited data or model misspecification \citep{Guo2017}.
For example, when using the continuous ranked probability score (CRPS), a widely used proper scoring rule, the resulting predictive distributions are often less dispersed than those obtained from maximum likelihood estimation \citep{buchweitz2025asymmetric,wessel2025enforcing}.


For univariate outcomes, probabilistic calibration builds on the probability integral transform (PIT) \citealt{Guo2017}. This diagnostic is particularly useful in practice, as systematic departures from uniformity not only indicate miscalibration but also reveal the nature of mis-specifications, if present. Moreover, several work have studied how calibration can be enforced during training in the univariate setting. \citet{Wilks2018} proposes augmenting the training objective based on scoring rules with an explicit penalty for miscalibration, thereby encouraging calibrated predictions. 

\begin{figure*}[!http]
  \centering
  \includegraphics[width=\textwidth]{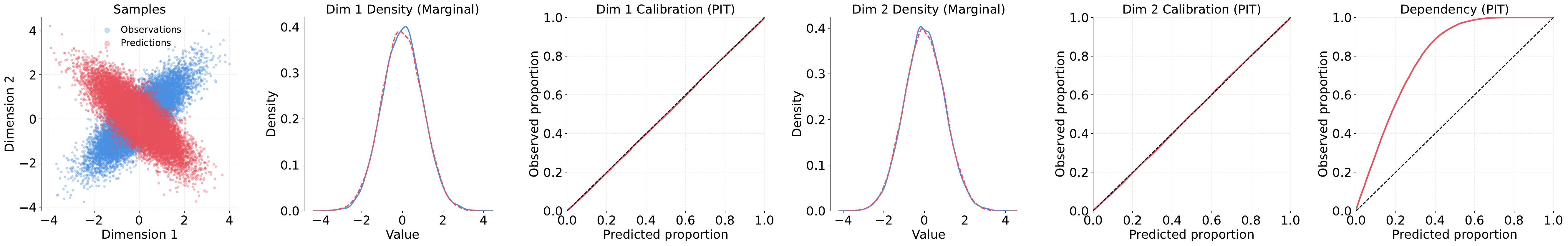}\\[2mm]
  \includegraphics[width=\textwidth]{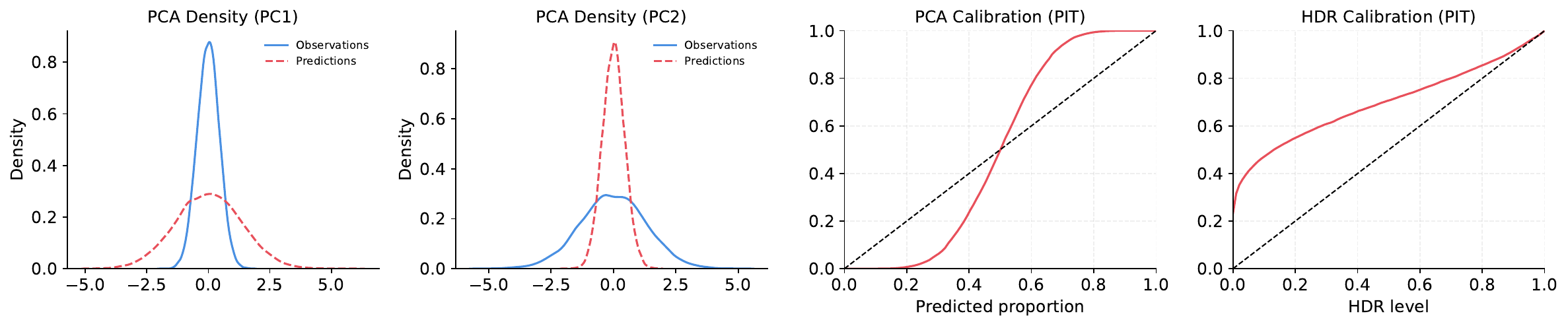}
  \caption{\textbf{Marginal calibration is insufficient for multivariate predictive distributions.}
  (Bottom) The model is well-calibrated marginally (per-dimension densities and PIT curves) yet fails to capture dependence, as shown by the mismatch in sample geometry and the strongly miscalibrated Dependency PIT plot.
  (Top) PCA-based diagnostics reveal miscalibration in projected subspaces, highlighting that calibration assessed only marginally can miss joint errors.}
  \label{fig:marginal_not_enough}
\end{figure*}

Although multivariate distributional regression models are widely used \citep{klein2024distributional}, calibration is still most often assessed only marginally, which does not ensure that the joint dependence structure is reliably calibrated, even when each marginal appears well calibrated \citep{Gneiting2008} (see figure~\ref{fig:marginal_not_enough} for an example).
Existing work has focused mainly on post-hoc diagnostics to assess multivariate calibration, most notably using multivariate rank histograms and their extensions \citep{Allen2023}. A pre-rank function maps a multivariate forecast-observation pair to a univariate summary that captures a specific aspect of forecast behavior. They may be chosen to assess different mis-specifications of the predictive distribution.

In this paper, we go beyond post-hoc diagnostics, and introduce a general framework that \textit{enforces} multivariate calibration into the training of distributional regression models. We further propose a \textit{new} pre-rank function based on principal components analysis (PCA), which assess calibration along the principal directions of the predictive distribution. Together, these contributions enable enforcement of multivariate calibration during training without sacrifing predictive performance.

Our main contributions are as follows:
\vspace{-0.3cm}
\begin{itemize}
    \item \textbf{Pre-rank Regularization.} We introduce a multivariate calibration framework for \textit{multivariate distributional regression} models by augmenting the training objective with a differentiable regularizer. The proposed regularization term penalizes deviations from uniformity of PIT values induced by a chosen pre-rank function, enabling calibration to be enforced during training.
    
    \item \textbf{PCA-based Pre-rank.} We propose a novel PCA-based pre-rank, which projects multivariate outcomes onto principal directions of the predictive distribution and is particularly sensitive to mis-specifications in the joint dependence structure. We further contrast it with other existing pre-ranks.

    \item \textbf{Empirical Evaluation.}  We conduct a simulation study to illustrate how different pre-ranks detect distinct forms of mis-specification, and perform a large-scale empirical evaluation on 18 real-world multi-output regression datasets to evaluate our methodology.

\end{itemize}

\section{Background}

We consider the problem of multivariate distributional regression, where the target variable $Y \in \mathcal{Y} \subseteq \mathbb{R}^D$ depends on an input variable $X \in \mathcal{X} \subseteq \mathbb{R}^L$.  We assume access to a training data $\mathcal{D} = \{(X_i, Y_i)\}_{i=1}^N,$ where $(X_i, Y_i) \overset{\text{i.i.d.}}{\sim} F_X \times F_{Y\mid X}$, where $F_{Y\mid X}$ is the true conditional distribution. 
We denote by $\hat F_{Y\mid X}^\theta$ a parametric predictive distribution with parameters $\theta$. When the dependence on $\theta$ is clear from context, we write simply $\hat F_{Y\mid X}$. Our objective is to approximate $F_{Y\mid X}$ using $\mathcal{D}$. Finally, we assume that predictive samples can be drawn from the estimated distribution
$\hat Y \mid X \sim \hat F_{Y\mid X}(\cdot \mid X).$

\paragraph{Proper scoring rules} 
Probabilistic regression models are commonly trained by minimizing a \emph{scoring rule}  \citep{winkler1996scoring}.  Let $\mathcal{P}(\mathcal{Y})$ denote a class of candidate predictive distributions over the outcome space $\mathcal{Y}$.

A scoring rule is a loss function  $S : \mathcal{P}(\mathcal{Y}) \times \mathcal{Y} \to \mathbb{R},$ which assigns a numerical score $S(\hat F, y)$ to a predictive distribution $\hat F \in \mathcal{P}(\mathcal{Y})$ when the outcome $y \in \mathcal{Y}$ is observed. 

A scoring rule is called \emph{proper} if, for any true distribution $F$,
\[
\mathbb{E}_{Y \sim F}\!\left[S(F,Y)\right] \le \mathbb{E}_{Y \sim F}\!\left[S(\hat{F},Y)\right]
\quad \text{for all }  \hat F \in \mathcal{P}(\mathcal{Y}),
\]
and \emph{strictly proper} if equality holds if and only if $\hat{F} = F$ \citep{GneitingRaftery2007, scheuerer2015variogram}.



\paragraph{Univariate calibration} In univariate regression ($D=1$), we recall probabilistic calibration \citep{Gneiting2007}. Let $\hat{F}_{Y \mid X}$ be a univariate predictive distribution. The associated PIT is defined as
\[
Z := \hat{F}_{Y \mid X}(Y \mid X),
\]

\begin{definition}[PIT calibration]
\normalfont
$\hat{F}_{Y \mid X}$ is \textit{probabilistically} calibrated if $Z\sim \mathcal{U}([0,1]) $, where $\mathcal{U}(0,1)$ is the uniform distribution.
\end{definition}

Under the true distribution, i.e.\ when $\hat{F}_{Y \mid X} = F_{Y \mid X}$, PIT calibration holds as a direct consequence of the PIT. In practice, deviations from uniformity of empirical PIT values provide a diagnostic of systematic forecast mis-specifications, such as bias or misrepresentation of dispersion \citep{Wilks2018}.


\paragraph{Probabilistic Calibration Error (PCE).}
\label{bac:pce}
In practice, calibration is assessed from a finite sample of PIT values $\{Z_i\}_{i=1}^N$, where $Z_i = \hat{F}_{Y\mid X=X_i}(Y_i)$.
A common approach is to compare the \textit{empirical} CDF
\begin{equation}
    \hat{F}_Z(\alpha) := \frac{1}{N}\sum_{i=1}^N \mathbf{1}\{Z_i \le \alpha\}, \qquad \alpha \in (0,1),
\end{equation}
to the CDF of the uniform distribution \citep{Gneiting2007}. A representative scalar discrepancy is the Probabilistic Calibration Error (PCE), defined on a fixed grid $\{\alpha_j\}_{j=1}^M \subset [0,1]$ as

\begin{equation}
\label{eq:PCE}
    \mathrm{PCE}(\hat{F}_{Y\mid X}) := \frac{1}{M}\sum_{j=1}^M \Big|\alpha_j - \hat{F}_Z(\alpha_j)\Big|,
\end{equation}
which measures the deviation of PIT empirical frequencies from the ideal uniform distribution. Such binning, or indicator-based estimators are effective for post-hoc evaluation, but are not directly suitable as training objectives due to non-differentiability and sensitivity to discretization \citep{Kumar2019, DheurBenTaieb2023}.

\subsection{Multivariate calibration}

When the target variable is multivariate ($D > 1$), defining and assessing probabilistic calibration becomes substantially more challenging. In particular, the multivariate PIT is no longer uniformly distributed, and multivariate quantiles are not uniquely defined, which prevents direct extensions of univariate PIT-based diagnostics \citep{GenestRivest2001,Gneiting2008}. Moreover, calibrating marginals alone is insufficient to ensure reliable joint predictions, as it does not involve the dependence structure of the predictive distribution \citep{ZiegelGneiting2014} (see Figure~\ref{fig:marginal_not_enough} for an example).


\paragraph{Assessing calibration via pre-rank functions.}

A general strategy to assess multivariate calibration is to reduce the multivariate problem to a collection of univariate ones by means of scalar projections. Following \citet{Allen2023}, we define calibration with respect to a pre-rank function.


\begin{definition}[Pre-rank function]
\normalfont
A \emph{pre-rank} is a function
\[
\rho : \mathcal{X} \times \mathcal{Y} \to \mathbb{R}
\]
that assigns a scalar summary to a forecast–observation pair $(X,Y)$. 

Given a pre-rank function $\rho$, we define the associated random variables
\begin{equation}
T := \rho(X,Y), 
\qquad 
\hat T := \rho(X,\hat Y),
\end{equation}
where $\hat Y$ denotes a sample drawn from the predictive distribution conditional on $X$.
\end{definition}

Since $\hat{T}$ is a scalar random variable, we define $\hat{F}_{T \mid X}$ induced by $\hat{F}_{Y \mid X}$. This allows us to define a \emph{projected PIT}
\begin{equation}
    Z_\rho := \hat{F}_{T \mid X}(T \mid X). \label{eq:zrho}
\end{equation}

\begin{definition}[Calibration with respect to a pre-rank]
\normalfont
A predictive distribution $\hat F_{Y\mid X}$ is said to be calibrated with respect to a pre-rank function $\rho$ if the projected PIT $Z_\rho$ is uniformly distributed on $[0,1]$.
\end{definition}

This formulation subsumes a wide range of existing multivariate calibration diagnostics, including copula-based calibration \citep{ZiegelGneiting2014} and density-based notions such as HDR calibration \citep{Chung2024}. Different choices of $\rho$ probe different aspects of the predictive distribution, such as location, scale, dependence structure, or tail behavior. Crucially, there are few restrictions on the choice of $\rho$, allowing practitioners to tailor calibration assessments to the aspect of interest.  Table~\ref{tab:preranks} gives examples of pre-rank functions.

\begin{table}[!htbp]
\centering
\small
\begin{tabular}{l|l}
\hline
Pre-rank & Formula \\
\hline
Marginal ($d$) &
$\rho_{\text{marg}}^d(x,y)=y_d$ \\

Location &
$\rho_{\text{loc}}(x,y)=\frac{1}{D}\sum_{d=1}^D y_d$ \\

Scale &
$\rho_{\text{scale}}(x,y)=\frac{1}{D}\sum_{d=1}^D (y_d-\bar y)^2$ \\

Dependency ($h$) &
$\rho_{\text{dep}}(x,y;h)=-\gamma_y(h)/s_y^2$ \\

HDR &
$\rho_{\text{hdr}}(x,y)=\hat f_{Y\mid X=x}(y)$ \\

Copula &
$\rho_{\text{cop}}(x,y)=\hat F_{Y\mid X=x}(y)$ \\
\hline
\end{tabular}

\caption{Pre-rank functions. Here $y=(y_1,\dots,y_D)\in\mathbb{R}^D$ denotes the multivariate target,
$\bar y=\frac{1}{D}\sum_{d=1}^D y_d$, $d\in\{1,\dots,D\}$ indexes the output dimensions,
$\gamma_y(h)=\frac{1}{2(D-h)}\sum_{d=1}^{D-h} |y_d-y_{d+h}|^2$ is a variogram-based dependency measure
with lag $h\in\{1,\dots,D-1\}$, and $s_y^2$ denotes the empirical variance across coordinates.}
\label{tab:preranks}
\end{table}

\section{A New Regularization-based Multivariate Calibration Method}

\label{sec:method}

In this section, we present a regularization-based calibration method that enforces multivariate calibration during the training of multivariate distributional regression models using pre-rank functions. We also introduce a novel PCA-based pre-rank and show how it integrates naturally within the proposed framework, alongside existing pre-ranks, to achieve different notions of multivariate calibration.



Following \citet{Wilks2018} and \citet{wessel2025enforcing}, who propose regularization-based calibration methods in the univariate setting, we aim to enforce calibration for multivariate predictions by augmenting the loss function with a regularization term based on a miscalibration measure. Specifically, our regularizer penalizes pre-rank miscalibration by measuring deviations from uniformity of the projected PIT values induced by a chosen pre-rank function.




In other words, if \( Z_\rho \) denotes the projected PIT associated with the pre-rank \( \rho \) as defined in \eqref{eq:zrho}, we define the regularizer as
\begin{equation}
\mathcal{R}(\hat{F}_{Y\mid X}; \rho)
= \frac{1}{M} \sum_{j=1}^M \left| \alpha_j - F_{Z_\rho}(\alpha_j) \right|
\label{eq:ideal-cal-reg}
\end{equation}
where $\{\alpha_j\}_{j=1}^M \subset [0,1]$ is a fixed grid of quantile levels. 

In practice, $F_{Z_\rho}$ is unknown and must be estimated from finite samples. A natural empirical estimator is given by the empirical CDF
\vspace{-0.25cm}
\begin{equation}
    \hat F_{Z_\rho}(\alpha) = \frac{1}{N} \sum_{i=1}^N \mathbf{1}\!\left(Z_{\rho,i} \le \alpha\right),
\end{equation}
where $Z_{\rho,i} = \hat{F}_{T_i \mid X=X_i}(T_i)$ are the projected PIT values computed on the training data and $T_i = \rho(X_i,Y_i)$. Substituting this estimator into \eqref{eq:ideal-cal-reg} yields and empirical PCE.




\paragraph{Differentiable Regularizer.} To obtain a differentiable training objective, we adopt the kernel-smoothed PCE (PCE-KDE) surrogate introduced by \citet{DheurBenTaieb2023}. Specifically, the indicator function is replaced by a smooth approximation using a logistic kernel, yielding the smoothed CDF
\begin{equation}
\Phi_{\text{KDE}}(\alpha_j; \{Z_{\rho,i}\}_{i=1}^N)
= \frac{1}{N} \sum_{i=1}^{N} \sigma\!\left( \tau(\alpha_j - Z_{\rho,i}) \right),
\label{eq:cdf-prerank-kde}
\end{equation}
where $\sigma(z) = \frac{1}{1 + e^{-z}}$ is the sigmoid function and $\tau > 0$ controls the smoothness of the approximation.

Replacing the empirical CDF in \eqref{eq:ideal-cal-reg} with its smoothed counterpart yields the final regularization term used in this work:
\begin{equation}
\begin{aligned}
\mathcal{R}_{\mathrm{KDE}}(\hat{F}_{Y\mid X},\rho)
&= \frac{1}{M}\sum_{j=1}^M
\Bigl|\alpha_j
- \Phi_{\mathrm{KDE}}\!\left(\alpha_j,\{Z_{\rho,i}\}_{i=1}^N\right)
\Bigr|^p,
\end{aligned}
\end{equation}


where $p \ge 1$ determines the shape of the penalty. 



\begin{figure*}[t]
    \centering
    \includegraphics[width=1\linewidth]{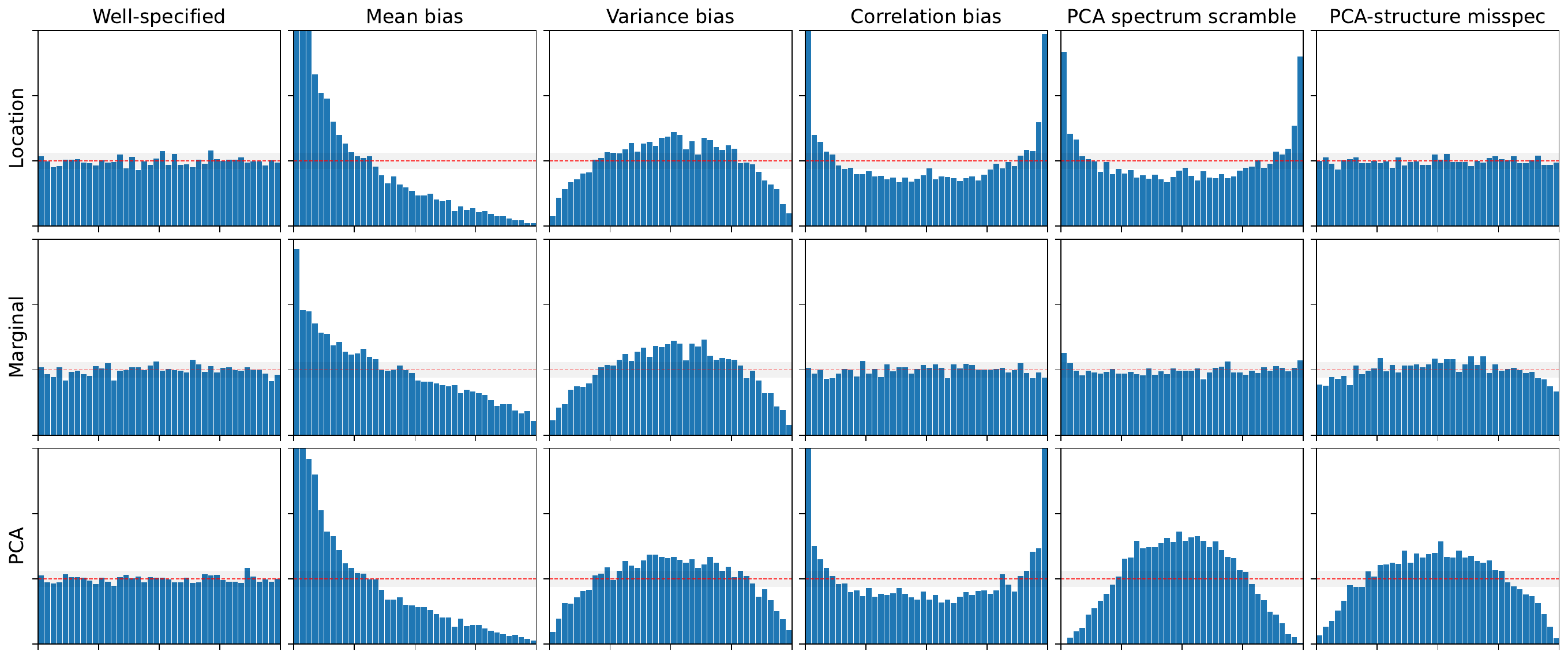}
    \caption{PIT histograms for mean, Marginal and PCA pre-rank under different examples of mis-specifications for a multivariate gaussian distribution. While the well-specified model exhibits an approximately uniform PIT, deviations from uniformity reveal errors in the distribution. In this example, the PCA pre-rank detects all considered forms of mis-specification. A full comparison across all pre-ranks is reported in Appendix~\ref{fig:Multivariate_Gaussian}--\ref{fig:Gaussian_Fields}.}
    \label{fig:pca_missp}
\end{figure*}


Let $S$ denote a strictly proper scoring rule used to train a multivariate distributional regression model with predictive distribution $\hat{F}_{Y\mid X}$. Given a pre-rank function $\rho$, we define the augmented objective function as

\vspace{-0.75cm}
\begin{equation}
\mathcal{L}
= \frac{1}{N} \sum_{i=1}^N
S\!\left(\hat{F}_{Y\mid X=X_i}, y_i\right)
+ \lambda \, \mathcal{R}_{\mathrm{KDE}}\!\left(\hat{F}_{Y\mid X = X_i}, \rho\right),
\label{eq:training-objective}
\end{equation}
where $\lambda \ge 0$ controls calibration enforcement. Setting $\lambda = 0$ in (\ref{eq:training-objective}) recovers standard unregularized training based solely on the scoring rule. The regularization term $\mathcal{R}_{\mathrm{KDE}}$ is differentiable through the projected PIT values, enabling end-to-end optimization using gradient-based methods. Algorithm~\ref{alg:train-prerank-kde} details the training process with the augmented loss.




\begin{algorithm}
\caption{Training with Pre-rank Regularization}
\label{alg:train-prerank-kde}
\begin{algorithmic}[1]
\STATE \textbf{Input:} data $\{(x_i,y_i)\}_{i=1}^N$, model $\hat F_{Y\mid X}^\theta$, scoring rule $S$, pre-rank $\rho$, $\lambda \ge0$, $M$, $\tau>0$, $p\ge1$, learning rate $\eta$, epochs $T$
\FOR{$t = 1,\dots,T$} 
  \FOR{minibatch $\mathcal{B}=\{(x_i,y_i)\}_{i=1}^B$}
    \STATE $T_i \gets \rho(x_i,y_i), \quad \forall i \in [B]$
    \STATE $\hat Y_i^{(m)} \sim \hat F_{Y\mid X=x_i}^{\theta_t}, \quad \forall i \in [B], m=1,\dots,M$
    \STATE $\hat T_i^{(m)} \gets \rho(x_i,\hat Y_i^{(m)})  \quad \forall i \in [B], m=1,\dots,M$
    \STATE $Z_{\rho,i} \gets \frac{1}{M}\sum_{m=1}^M \mathbf{1}\{\hat T_i^{(m)} \le T_i\}, \quad \forall i \in [B]$
    \FOR{$j=1,\dots,M$}
      \STATE $\Phi_{\mathrm{KDE}}(\alpha_j) \gets \frac{1}{B}\sum_{i=1}^B \sigma\!\big(\tau(\alpha_j - Z_{\rho,i})\big)$
    \ENDFOR
    \STATE $\mathcal{R}_{\mathrm{PCE\text{-}KDE}} \gets \frac{1}{M}\sum_{j=1}^M \big|\alpha_j - \Phi_{\mathrm{KDE}}(\alpha_j)\big|^p$
    \STATE $\mathcal{L} \gets \frac{1}{B}\sum_{i=1}^B S(\hat F_{Y\mid X=x_i}^{\theta_t},y_i) + \lambda\,\mathcal{R}_{\mathrm{PCE\text{-}KDE}}$
    \STATE $\theta_{t+1} \gets \theta_t - \eta\nabla_{\theta_t} \mathcal{L}$
  \ENDFOR
\ENDFOR
\STATE \textbf{Return:} $\hat F_{Y\mid X}^{\theta_T}$
\end{algorithmic}
\end{algorithm}


\paragraph{Different Pre-rank Functions.} Our framework is agnostic to the choice of pre-rank function. In this work, we include a diverse set of pre-ranks proposed in the literature \citep{scheuerer2015variogram, knuppel2022score, Allen2023}, each targeting a specific aspect of multivariate miscalibration. Specifically, we consider marginal, location, scale, dependence-based, copula-based \citep{ZiegelGneiting2014}, and high-density region (HDR) pre-ranks \citep{Chung2024}. While copula- and HDR-based recalibration are post-hoc methods, we adapt them to our training-time, regularization-based calibration framework by defining their respective pre-rank functions as listed in Table~\ref{tab:preranks}.





\begin{algorithm}[t]
\caption{PCA Pre-Rank $\rho^{\mathrm{PCA}}_k(x,y)$}
\label{alg:pca-prerank}
\begin{algorithmic}[1]


\STATE \textbf{Input:} Observation $y \in \mathbb{R}^D$, samples $\{\hat{Y}^{(m)}\}_{m=1}^M$ with $\hat{Y}^{(m)} \sim \hat{F}_{Y\mid X=x}$, index $k \le \min(D, M-1)$



\STATE Compute the covariance of $\{\hat Y^{(m)}\}_{m=1}^M$
\STATE Compute the $k$-th principal component $u_k$ 
\STATE $T_{k}\gets \langle y,u_k\rangle$
\STATE \textbf{Output:} $T_{k}$
\end{algorithmic}
\end{algorithm}

\section{A New PCA-based Pre-rank}

We introduce a new pre-rank function based on principal component analysis (PCA) that aims to detect miscalibration along the principal directions of variation of the predictive distribution. Existing pre-ranks either target a single aspect of the predictive distribution or require several projections that grows linearly with the output dimension, as in marginal pre-rank.


We assess calibration along principal directions because they correspond to projections with maximal predictive variance, where uncertainty is largest and calibration errors are most detectable, random directions typically probe low-variance averages and are less sensitive to misspecifications.


For a pair $(x, y)$, we draw samples from $\hat{F}_{Y\mid X=x}$ and estimate the associated covariance matrix $\hat\Sigma_x$. If $v_k(x)$ is the associated $k$-th principal component, the $k$-th PCA pre-rank is given by
\begin{equation}
    \rho^{\mathrm{PCA}}_k(x,y) = v_k(x)^\top y.
\end{equation}

Algorithm~\ref{alg:pca-prerank} summarizes the steps to compute the $k$-th PCA pre-rank.

\paragraph{Miscalibration detection with the PCA Pre-rank.} 

From a ranking perspective, PCA satisfies several desirable invariance and symmetry properties. It is equivariant under centering, linear re-scaling, and orthogonal transformations, and is permutation-equivariant with respect to coordinate relabeling. As a covariance-based method, PCA depends only on second-order structure, which yields a computationally stable diagnostic in high-dimensional settings.

Unlike other linear projections, PCA is optimal in maximizing projected variance \citep{jolliffe2002principal}. Consequently, the leading principal components concentrate directions along which covariance misspecification is most pronounced, making them particularly informative for calibration assessment under second-order model errors.

The role of the PCA pre-rank is illustrated through a simulation study in Section~\ref{sec:experiments}. We compare PCA pre-rank to existing approaches, including mean, marginal, density-based (HDR), dependence-based, and variance pre-ranks, across a range of misspecifications such as mean and variance bias, correlation bias, and perturbations that alter the eigenstructure of the predictive covariance while preserving marginal distributions. PCA pre-rank systematically yields non-uniform projected PIT distributions under such misspecifications, compared to other pre-ranks, as shown in Figure~\ref{fig:pca_missp}.

\begin{figure*}[!htbp]
    \centering
    \begin{subfigure}[b]{0.24\linewidth}
        \includegraphics[width=\linewidth]{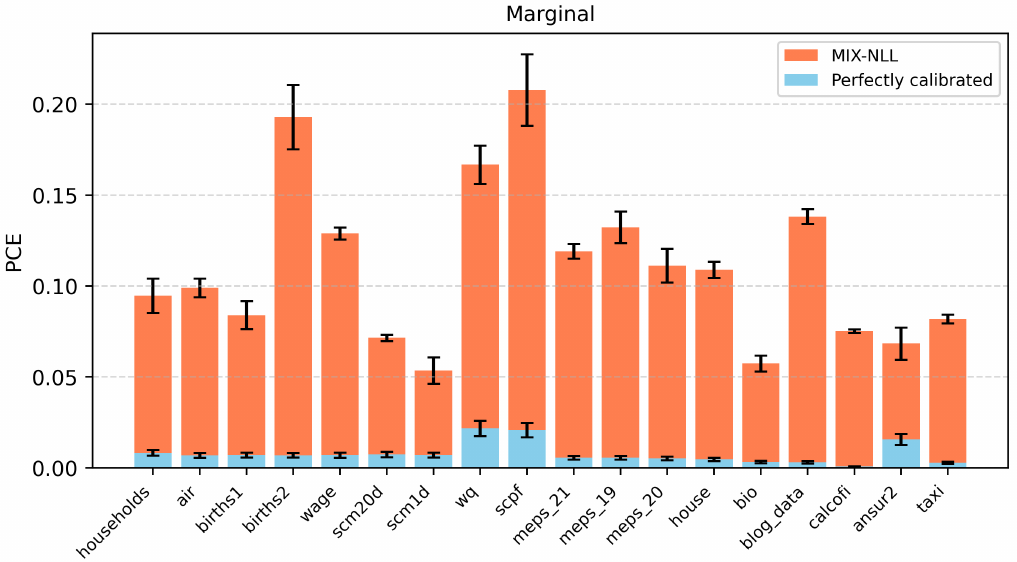}
        \caption{Marginal}
        \label{subfig:marg}
    \end{subfigure}
    \begin{subfigure}[b]{0.24\linewidth}
        \includegraphics[width=\linewidth]{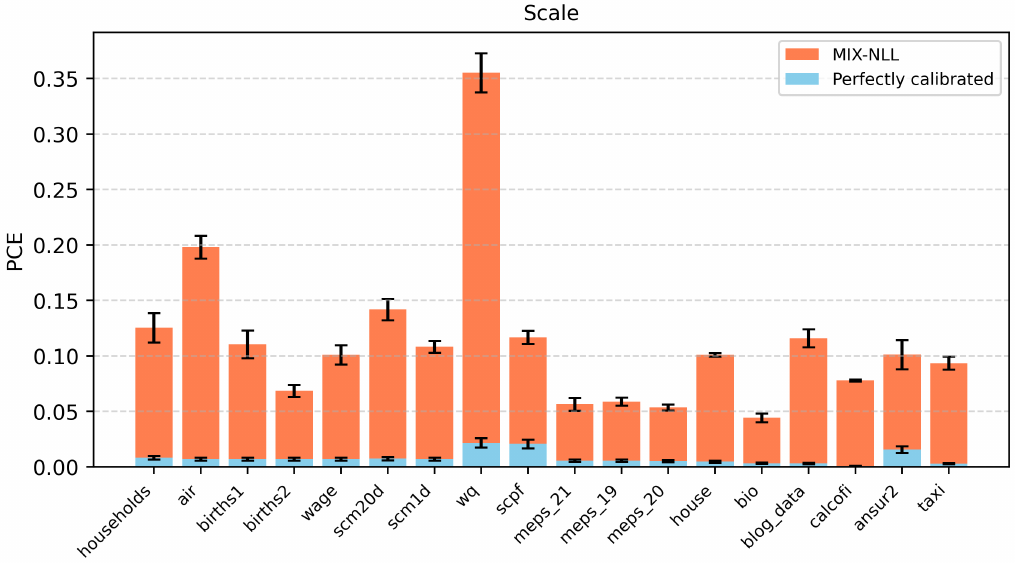}
        \caption{Scale}
        \label{subfig:var}
    \end{subfigure}
    \begin{subfigure}[b]{0.24\linewidth}
        \includegraphics[width=\linewidth]{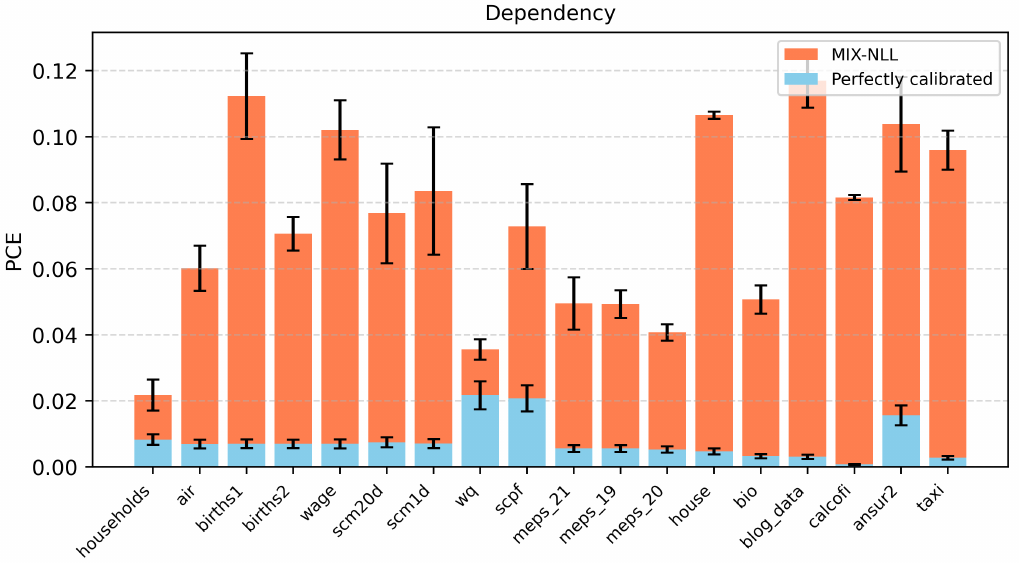}
        \caption{Dependency}
        \label{subfig:dep}
    \end{subfigure}
    \begin{subfigure}[b]{0.24\linewidth}
        \includegraphics[width=\linewidth]{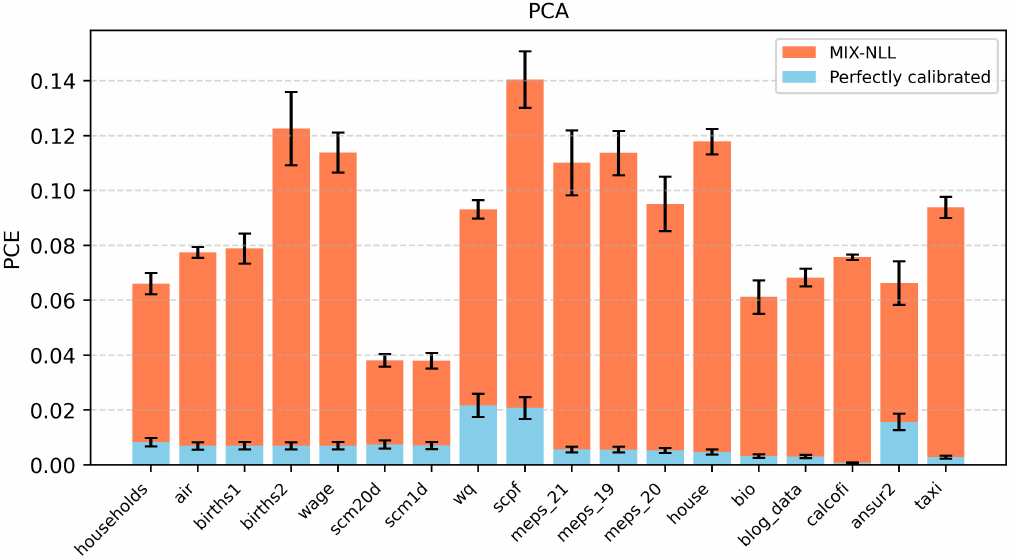}
        \caption{PCA}
        \label{subfig:pca}
    \end{subfigure}
    \caption{PCE values with respect to (a) Marginal (b) Scale (c) Dependency and (d) PCA pre-ranks averaged over five runs across 18 benchmark datasets using the MIX-NLL baseline. Blue bars indicate reference PCE values from a simulated perfectly calibrated model (defined as the oracle predictor satisfying $\hat{F}_{Y \mid X} = F_{Y \mid X}$).}
    \label{fig:hyp_test_therest}
\end{figure*}

\section{Experiments}
\label{sec:experiments}

We revisit the diagnostic framework and simulation studies by \citet{thorarinsdottir2018verification, Allen2023}. We first study the sensitivity of different pre-ranks under controlled misspecifications, including under- and over-estimation of the mean, variance $\sigma^2$, correlation $\tau$, as well as PCA-targeted covariance perturbations (see more details about misspecifications in Appendix~\ref{app:cov-misspec}). These simulations illustrate which forms of miscalibration are exposed by existing pre-ranks and motivate the use of our proposed PCA pre-rank for detecting dependence-structure misspecifications that remain invisible to marginal or lower-order diagnostics. 

In the second part of this section, we evaluate the practical impact of our multivariate regularization framework on a suite of real-world regression datasets, assessing calibration improvements alongside predictive performance.

The experiments are designed to address three main questions: (i) which forms of multivariate miscalibration are detected by different pre-rank functions, (ii) whether enforcing calibration via pre-rank regularization improves calibration without degrading predictive performance, and (iii) whether the proposed PCA-based pre-rank captures calibration failures that are not detected by existing pre-ranks.

Figure \ref{fig:hyp_test_therest} shows PCE values for the marginal, scale, dependency, and PCA pre-ranks, averaged over five independent runs on 18 real benchmark datasets using the MIX-NLL model. Simulated PCE scores from a perfectly calibrated model are shown in blue.

\subsection{Simulation study}
\label{sec:simulations}
We consider different scenarios to examine the sensitivity of pre-rank functions to different forms of mis-specification. In all simulations, calibration is assessed using projected PIT histograms. For a well-calibrated predictive distribution, the PIT is uniformly distributed on $[0,1]$, yielding a flat histogram. Systematic deviations from uniformity indicate specific forms of miscalibration: U-shaped histograms reflect underdispersion, inverted-U shapes indicate overdispersion, and skewness signals bias in the projected distribution.

\subsubsection{Simulation 1 (Multivariate Gaussian).}
\label{sec:sim1}
The first example is a standard multivariate gaussian distribution. Observations are generated as $Y \sim \mathcal{N}(0, \Sigma)$ where $\Sigma \in \mathbb{R}^{D\times D}$ is the covariance matrix defined by :  $\Sigma_{ij} = \sigma^2 \exp(\frac{-|i-j|}{\ell}), \quad \ell = 1,$ with $d=10$ and $\sigma =1$. For each observation, $M=20$ ensemble members are drawn at random from a mis-specified multivariate
normal distribution.

We consider a collection of mis-specifications affecting different aspects of the predictive distribution, summarized in Table~\ref{tab:sim1_misspec} in Appendix~\ref{app:simulation}. These include classical perturbations of the mean, variance, and correlation, as well as covariance perturbations that preserve marginal distributions but alter the eigenstructure of $\hat \Sigma$. 

Figure~\ref{fig:Multivariate_Gaussian} reports the projected PIT histograms for different pre-rank functions under the mis-specifications listed in Table~\ref{tab:sim1_misspec}. Location pre-rank is sensitive to most mis-specifications that affect the predictive distribution (mean, variance and correlation), but fail to detect the PCA-structure mis-specification. Marginal and HDR pre-ranks detect biases in the mean and variance, but remain largely insensitive to mis-specifications that affect the joint covariance structure. Dependency pre-rank is insensitive to pure mean and variance errors, yet respond strongly to covariance perturbations that modify principal directions. The scale pre-rank is, as expected, highly sensitive to variance bias and PCA-based errors while remaining insensitive to pure mean shifts. In the other hand, the PCA pre-rank, as shown in Figure~\ref{fig:pca_missp}, consistently detects all forms of mis-specifications considered in this simulation, including those that alter the covariance eigenstructure, a comparison with all other pre-ranks is shown in Figure~\ref{fig:Multivariate_Gaussian} in Appendix~\ref{app:simulation}.

\begin{figure*}[t]
    \centering
    \begin{subfigure}[b]{0.33\linewidth}
        \includegraphics[width=\linewidth]{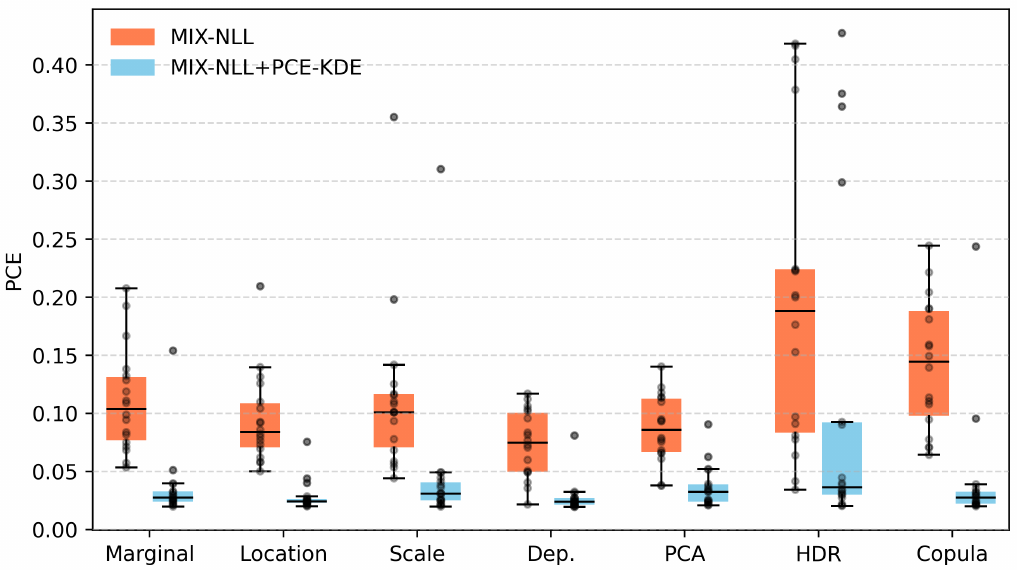}
        \caption{PCE}
        \label{subfig:pce}
    \end{subfigure}
    \begin{subfigure}[b]{0.33\linewidth}
        \includegraphics[width=\linewidth]{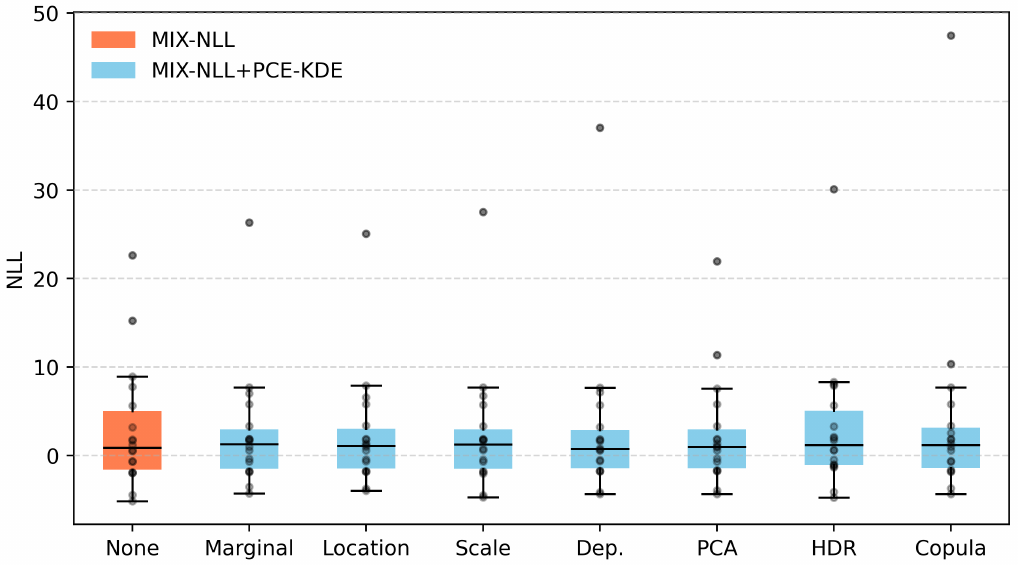}
        \caption{NLL}
        \label{subfig:nll}
    \end{subfigure}
    \begin{subfigure}[b]{0.33\linewidth}
        \includegraphics[width=\linewidth]{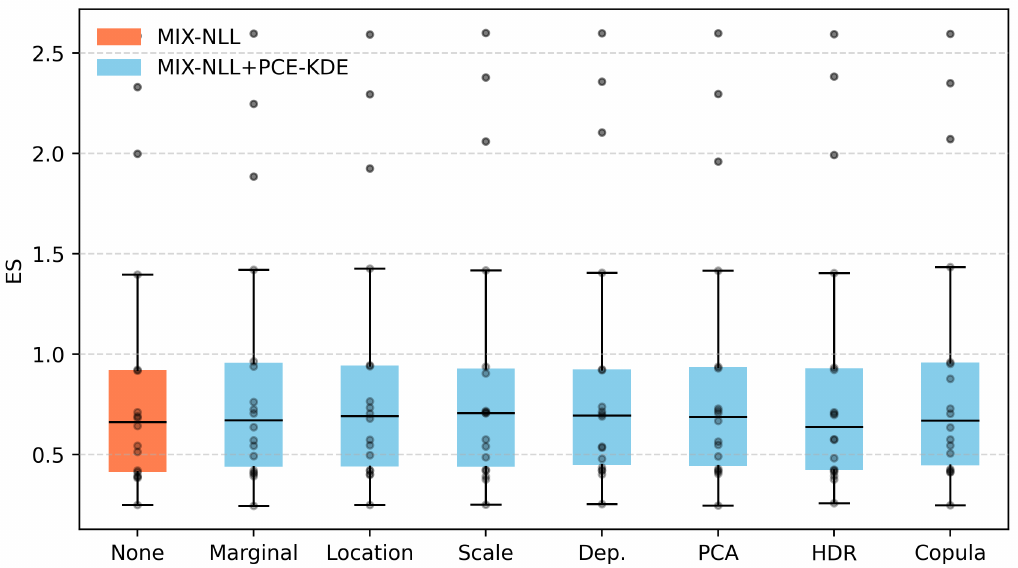}
        \caption{Energy Score}
        \label{subfig:es}
    \end{subfigure}
    \caption{Performance on benchmark of 18 datasets. Orange: MIX-NLL (no regularization). Blue: \texttt{MIX-NLL+PCE-KDE} (our proposed model). Metrics are calculated across seven pre-rank functions, and averaged over five runs. In subplots (b) and (c), the “None” box refers to the unregularized \texttt{MIX-NLL} trained without pre-rank.}
    \label{fig:metrics-before-after}
\end{figure*}

\subsubsection{Simulation 2 (Gaussian random fields)}
\label{sec:sim2}
We next consider a higher-dimensional and spatially structured example to assess whether the patterns observed in simulation 1 persist in the presence of spatial dependence in the data. Observations are generated as realizations of a gaussian random field on a $5 \times 5$ grid, $Y \sim \mathcal{N}(0, \Sigma),$ where $\Sigma \in \mathbb{R}^{d \times d}$ with $d=25$ is defined by an isotropic exponential covariance function based on Euclidean distance $ \Sigma_{ij} = \sigma^2 \exp\!\left(-\frac{\|s_i - s_j\|}{\ell}\right), \quad \ell = 1,$ and $s_i$ denotes the spatial location of grid point $i$. 
 
These miscalibration scenarios induce substantial errors in the joint dependence structure. 

Figure~\ref{fig:Gaussian_Fields} in Appendix~\ref{app:simulation} reports the projected PIT histograms under these mis-specifications. The qualitative behavior closely mirrors that observed in simulation 1. Marginal and location pre-ranks primarily respond to biases in the mean or marginal variances, but fail to detect mis-specifications that affect spatial dependence or covariance geometry. Dependency-based pre-rank is sensitive to changes in spatial correlation and anisotropy, but does not detect pure mean or variance biases. PCA pre-rank remain sensitive across all considered scenarios, including anisotropic and covariance perturbations that leave marginal behavior unchanged. 

Overall, simulations 1 and 2 illustrate that the PCA pre-rank is sensitive to joint miscalibration in multivariate settings, revealing mean, variance, and correlation errors as well as dependence-structure distortions in cases where existing pre-rank diagnostics are insufficient. These observations motivate the integration of the PCA pre-rank into our framework, alongside the previously introduced pre-ranks.


\subsection{Experiments with Real-world Data}

\subsubsection{Data}
We conduct our experiments on 18 datasets drawn from prior work \citep{Guan2021LocalizedCP, feldman2022, wang2022probabilisticconformalpredictionusing, DheurBenTaieb2023, camehl2025, Dheur2025}. They serve as a standard benchmark for evaluating calibration methods. We include only datasets with at least 400 training instances and follow the same preprocessing and train-validation-test splitting procedure as in \citet{Dheur2025}. The selected datasets vary in size, containing between 424 and 406,440 training examples. The number of input features $L$ ranges from 1 to 279, and the number of output variables $D$ ranges from 2 to 16.

Our base multivariate distributional regression model is a mixture of $K$ multivariate Gaussian components, where all parameters are generated by a hypernetwork (details in Appendix~\ref{app:model_hyper}). This model is then trained with our multivariate calibration framework, using a strictly proper scoring rule, negative log likelihood (NLL) \citep{winkler1996scoring}, augmented with our proposed regularizer, following the training procedure outlined in Algorithm~\ref{alg:train-prerank-kde}. Calibration is then evaluated using PCE as defined in (\ref{bac:pce}). Predictive performance is assessed using both NLL and energy score (ES).

For each pre-rank, projected PIT values are estimated using samples from the predictive distribution. The regularization weight $\lambda$ is selected on a validation set to balance calibration and predictive performance, we report the tuning results in table~\ref{tab:lambda-tuning} in Appendix~\ref{app:model_hyper}. All results are averaged over 5 random seeds.

To contextualize observed PCE values, we report, in Figure~\ref{fig:hyp_test_therest}, reference levels obtained from a simulated perfectly calibrated model. These values represent the null behavior of the PCE under ideal calibration.

\begin{figure*}[t]
    \centering
    \includegraphics[width=1\linewidth]{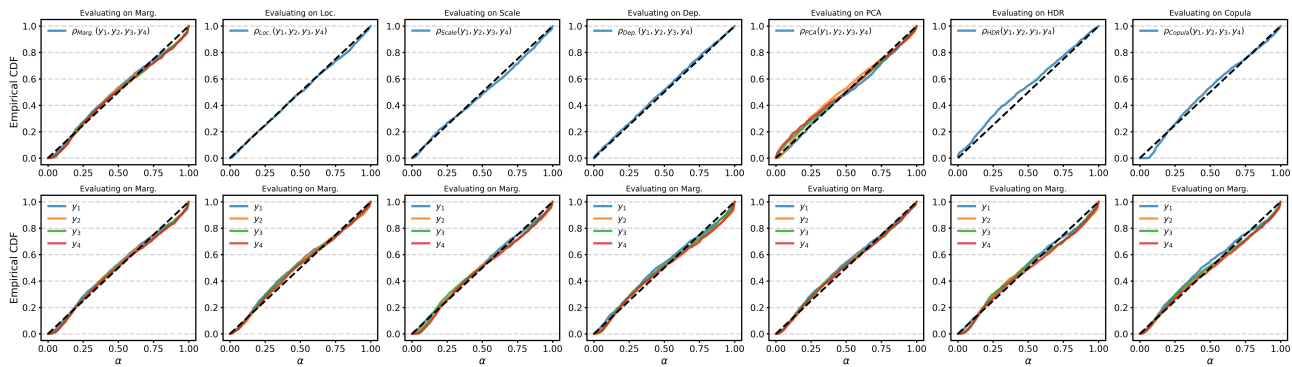}
    \caption{Reliability plots on the \texttt{households} dataset using the \texttt{MIX-NLL+PCE-KDE} model after pre-rank regularization. The top row shows calibration curves evaluated with respect to different pre-ranks, while the bottom row reports the corresponding marginal calibration curves.}

    \label{fig:housereli}
\end{figure*}

\subsubsection{Results}

Figure~\ref{subfig:pce} compares the PCE of the unregularized baseline model (\texttt{MIX-NLL}) with that of models trained using our pre-rank regularization framework (\texttt{MIX-NLL+PCE-KDE}). As expected, regularization leads to a substantial reduction in PCE for the corresponding pre-rank. Moreover, the distribution of PCE values across datasets, visualized via box plots, is markedly tighter, indicating more stable and consistent calibration improvements.
Complete results, averaged over five independent runs for each dataset and pre-rank, are reported in Appendix~\ref{app:results}.
Figures~\ref{subfig:nll} and~\ref{subfig:es} report the corresponding NLL and ES, comparing models trained with pre-rank regularization to the unregularized baseline (denoted ``None''). Across all settings, we observe no systematic degradation in predictive performance. This suggests that enforcing multivariate calibration through pre-rank regularization can be achieved without sacrificing predictive performance.

\paragraph{Flexibility and Complementarity of Pre-ranks.}
As emphasized by \citealt{Allen2023}, pre-rank functions offer flexibility for probing different aspects of multivariate calibration. Each pre-rank targets a specific property of the predictive distribution, such as location, scale, or dependence, allowing practitioners to focus on the calibration aspects most relevant to their application.

\vspace{-0.5cm}
\paragraph{Role of the PCA-based Pre-rank.}
Principal components identify the directions along which the predictive distribution exhibits the largest variability. By assessing calibration along these principal components of the predictive covariance, the proposed PCA pre-rank focuses calibration on directions that explain the largest predictive variance, thereby capturing the primary dependence structure of the learned predictive distribution. Combined with the simulation study, these results show that the PCA pre-rank captures calibration errors that are not detected by existing pre-ranks, thereby providing a complementary view of multivariate calibration.

More broadly, our framework supports multiple forms of multivariate calibration through the choice of pre-rank, rather than committing to a single diagnostic perspective.

\vspace{-0.3cm}
\paragraph{Effectiveness of Pre-rank regularization.}
The effectiveness of the proposed calibration framework is further illustrated through empirical experiments. Figure~\ref{fig:housereli} report reliability plots for \texttt{households} dataset, comparing different pre-ranks. Across all cases, the regularized models exhibit reliability curves that more closely align with the diagonal, indicating improved calibration between predicted and observed probabilities.

Complementing these diagnostics, Figure~\ref{fig:metrics-before-after} summarizes results across all 18 benchmark datasets. Pre-rank regularization consistently reduces the PCE across pre-ranks, while preserving predictive accuracy as measured by the NLL and energy score. Table~\ref{tab:real-pce-good-1col-top5} focuses on a representative subset of calibrated datasets, and highlights reductions in PCE relative to the unregularized baseline across all pre-ranks.

\section{Conclusion}

We introduce a new multivariate calibration framework for multivariate distributional regression by augmenting the training objective with a differentiable pre-rank based regularizer. This approach improved calibration by penalizing deviations from uniformity of projected PIT values with respect to flexible pre-ranks.

Within this framework, we proposed a \textit{novel} PCA-based pre-rank designed to assess calibration along principal directions explaining the largest predictive variance, and adapted existing post-hoc calibration methods, specifically HDR, and Copula recalibrations, into training-time pre-rank functions. Simulation results demonstrate that the PCA pre-rank detects dependence-structure misspecifications that remain invisible to existing diagnostics.

Across 18 real-world multi-output regression datasets, pre-rank regularization consistently improves multivariate calibration without degrading predictive performance. Overall, our results show that multivariate calibration can be enforced during training, with the choice of pre-rank determining which forms of miscalibration are addressed. Future investigations may extend this method to other models, and explore alternative notions of calibration.





\begin{table}[t]
\centering
\scriptsize
\begin{tabular}{lcccccc}

\toprule
Pre-rank & bio & calcofi & taxi & blogdata & meps21 & meps19 \\
\midrule
\textit{None} & 0.061 & 0.076 & 0.094 & 0.068 & 0.110 & 0.114 \\
Marg.         & 0.021 & 0.020 & 0.021 & 0.023 & 0.026 & 0.026 \\
Loc.          & 0.020 & 0.021 & 0.022 & 0.024 & 0.025 & 0.023 \\
Scale         & 0.021 & 0.020 & 0.025 & 0.023 & 0.031 & 0.050 \\
Dep.          & 0.021 & 0.020 & 0.023 & 0.023 & 0.024 & 0.025 \\
PCA           & 0.021 & 0.021 & 0.022 & 0.025 & 0.025 & 0.024 \\
HDR           & 0.021 & 0.020 & 0.022 & 0.030 & 0.032 & 0.031 \\
Copula        & 0.021 & 0.020 & 0.022 & 0.024 & 0.023 & 0.022 \\
\bottomrule
\end{tabular}
\caption{PCE values for five datasets, reported across all pre-rank functions and compared to the unregularized baseline (\textit{None}). Full tables~\ref{tab:pce_before_formatted} and \ref{tab:real-pce-after} in Appendix.}
\label{tab:real-pce-good-1col-top5}
\end{table}

\newpage

\bibliography{references}
\bibliographystyle{icml2026}

\newpage
\onecolumn
\appendix
\section{Related Work}
\label{app:relatedwork}
Calibration in probabilistic regression has been studied extensively in the univariate setting \citep{Dawid1984,Diebold1998,Gneiting2007}. While strictly proper scoring rules such as the negative log-likelihood (NLL) or the continuous ranked probability score (CRPS) are widely used for training probabilistic models \citep{winkler1996scoring,GneitingRaftery2007}, they do not guarantee calibrated predictions. This has motivated methods that explicitly encourage calibration during training. \citet{Wilks2018} introduces calibration penalties for ensemble post-processing, and more recently, \citet{wessel2025enforcing} propose training-time regularization schemes that enforce calibration with respect to specific aspects of the predictive distribution, such as tail behavior. Related work has also explored regularization and loss-based approaches to improve calibration in univariate regression \citep{gebetsberger2018estimation,buchweitz2025asymmetric}. These methods, however, are fundamentally designed for scalar targets and do not address calibration of joint predictive distributions for multivariate outcomes.

Extending calibration notions to multivariate regression is considerably more challenging due to the need to account for dependence structures across target dimensions. Early work by \citet{Gneiting2008} introduces multivariate rank histograms as diagnostic tools for assessing probabilistic calibration of multivariate forecasts. Copula-based notions of calibration further formalize joint calibration by evaluating the uniformity of the copula probability integral transform \citep{ZiegelGneiting2014}. More recently, \citet{Allen2023} propose a unifying diagnostic framework based on pre-rank functions. Their work demonstrates that different pre-ranks probe complementary aspects of multivariate miscalibration, such as location, scale, or dependence, but focuses exclusively on evaluation rather than enforcement. Futher approaches aim to improve multivariate calibration via post-hoc recalibration. Copula-based recalibration methods adjust joint distributions after training, but typically require access to the full joint CDF and are not easily integrated into gradient-based learning pipelines. Sampling-based methods such as HDR recalibration \citep{Chung2024} leverage density-level projections to recalibrate predictive samples with respect to highest density regions, yielding joint calibration guarantees in a post-processing step. While effective in some settings, these approaches depend critically on the quality of the estimated predictive density and are not designed to influence the training dynamics of the underlying model.

\section{Additional details}

\paragraph{Practical note on Copula-based Pre-Rank.}
When using copula-based pre-ranks, one requires access to the joint CDF \( \hat{F}_{Y|X}(y) \), i.e., the probability that all components of \( Y \) are less than or equal to \( y \) given \( X \). However, many models provide only the conditional density \( \hat{f}_{Y|X} \).

We approximate the joint CDF via Monte Carlo sampling. Given input \( X_i \) and target \( Y_i \), we draw \( S \) samples \( \hat{Y}_{i,1}, \dots, \hat{Y}_{i,S} \sim \hat{f}_{Y|X=X_i} \) and estimate:
\begin{equation}
\hat{F}_{Y|X=X_i}(Y_i) \;\approx\; \frac{1}{S} \sum_{s=1}^S \mathbf{1}\left\{ \hat{Y}_{i,s} \leq Y_i \right\},
\end{equation}
where the indicator \( \mathbf{1}\left\{ \hat{Y}_{i,s} \leq y_i \right\} \) is true if and only if \( \hat{Y}_{i,s}^{(d)} \leq y_i^{(d)} \) for all components \( d = 1, \dots, D \).

Since the indicator function is not differentiable, we replace it with a smooth approximate using the sigmoid function \( \sigma(z) = 1/(1 + e^{-z}) \) and a temperature parameter \( \tau > 0 \). This gives:
\begin{equation}
\hat{F}_{Y|X=X_i}(Y_i) \;\approx\; \frac{1}{S} \sum_{s=1}^S \prod_{d=1}^{D} \sigma\left( \tau \Big(Y_i^{(d)} - \hat{Y}_{i,s}^{(d)}\Big) \right),
\end{equation}
where \( y_i^{(d)} \) and \( \hat{Y}_{i,s}^{(d)} \) denote the \( d \)-th components of the vectors \( y_i \) and \( \hat{Y}_{i,s} \), respectively. The product over dimensions enforces that all components of \( \hat{Y}_{i,s} \) fall below the threshold \( y_i \), mimicking the joint indicator condition.

This smooth approximation is fully differentiable with respect to the model parameters (via the samples \( \hat{Y}_{i,s} \)), and thus compatible with gradient-based optimization routines such as backpropagation.

\paragraph{Empirical Calculation of Energy Score}
We use Energy Score (ES) as a scoring rule metric to evaluate our model performance. ES generalizes Continuous Ranked Probability Score (CRPS) to multivariate settings and is computed empirically as:
\begin{equation}
    \text{ES}(\hat{F}, y) = \frac{1}{G} \sum_{i=1}^G \|\hat{Y}_i - y\| - \frac{1}{2G^2} \sum_{i=1}^G \sum_{j=1}^G \|\hat{Y}_i - \hat{Y}_j\|
\end{equation}
where $\{\hat{Y}_i\}_{i=1}^G \sim \hat{F}_{Y|X}$ are $G$ samples drawn from the predictive distribution. We set $G=100$ in all experiments.

\subsection{Pre-rank functions}

\paragraph{Equivalence of Pre-Rank Calibration.}
\begin{proposition}
\emph{For every fixed $x \in \mathbb{R}^L$, the function $y \mapsto \rho_2(x, y)$ must be a strictly monotonic bijective transformation of $y \mapsto \rho_1(x, y)$. That is, there exists a strictly increasing or decreasing bijection $h_x$ such that for all $y \in \mathbb{R}^D$,}
$$
\rho_2(x, y) = h_x(\rho_1(x, y)).
$$
\end{proposition}
\textit{Proof.} 
Fix $x \in \mathbb{R}^L$ and define $T_1 = \rho_1(x, Y)$ and $T_2 = \rho_2(x, Y) = h_x(T_1)$, where $h_x$ is a strictly monotonic bijection. Let $\hat{F}_{T_1 \mid X= x}$ and $\hat{F}_{T_2 \mid X= x}$ denote the empirical conditional CDFs of $T_1$ and $T_2$, respectively, estimated using the same sample of predicted values $\{\hat{Y}_i\}_{i=1}^{N'}$ drawn from the predictive distribution $\hat{F}_{Y \mid X= x}$.

As explained in the background section, we estimate these conditional CDFs using the empirical estimator. Since this construction is used solely for evaluation, differentiability of the CDF is not required. Then for any $t \in \mathbb{R}$,
\begin{align*}
\hat{F}_{T_2 \mid X= x}(t) 
&=  \frac{1}{N'} \sum_{i=1}^{N'} \mathbf{1}_\tau(\rho_2(x,\hat{Y}_i) \leq t)\\
&=  \frac{1}{N'} \sum_{i=1}^{N'} \mathbf{1}_\tau(\rho_1(x,\hat{Y}_i) \leq h_x^{-1}(t))\\
&=  \hat{F}_{T_1 \mid X= x}(h_x^{-1}(t)).
\end{align*}

Since \( T_2 = h_x(T_1) \) and 
\[
\hat{F}_{T_2 \mid X= x}(t) = \hat{F}_{T_1 \mid X= x}(h_x^{-1}(t)),
\] we have:
\[
\hat{F}_{T_2 \mid X= x}(T_2) = \hat{F}_{T_1 \mid X= x}(h_x^{-1}(T_2)) = \hat{F}_{T_1 \mid X= x}(T_1),
\] where we used the fact that \( T_1 = h_x^{-1}(T_2) \) by construction. It follows that the PIT value computed under \( \rho_2 \) coincides with the one computed under \( \rho_1 \):
\[
U_2 := \hat{F}_{Z_2 \mid X= x}(Z_2) = \hat{F}_{Z_1 \mid X= x}(Z_1) =: U_1.
\] It follows that \( U_1 \) and \( U_2 \) have the same distribution. In particular,
\[
U_2 \sim \mathcal{U}[0,1] \quad \Longleftrightarrow \quad U_1 \sim \mathcal{U}[0,1],
\]
which establishes the equivalence of the two calibration criteria under the assumed transformation.

\paragraph{Flexibility in the choice of pre-ranks.}
As emphasized by \citealt{Allen2023}, there is substantial flexibility in the choice of pre-rank functions. Different pre-ranks probe different aspects of the predictive distribution, allowing practitioners to select them on a case-by-case basis depending on which calibration properties are most relevant for their application. In some settings, accurate variance estimation may be critical, in others, capturing joint dependence structures is essential, and in simpler cases, assessing the calibration of the mean alone may be sufficient.

To illustrate what each pre-rank can and cannot detect under different forms of distributional misspecification, we conduct a simulation study. The resulting PIT histograms, reported in Figures~\ref{sec:sim1} and~\ref{sec:sim2} in Appendix~\ref{app:simulation}, are straightforward to interpret and make explicit which systematic deficiencies a given pre-rank is sensitive to. Deviations from uniformity directly reveal specific forms of miscalibration, such as biases in location, dispersion, or dependence structure, thereby facilitating a clear and interpretable assessment of predictive quality.

\section{Simulation studies}
\label{app:simulation}
\paragraph{Simulation 1 (Multivariate Gaussian).}
We first study a synthetic $d=10$ dimensional multivariate Gaussian setting in which the true data-generating distribution is $y \sim \mathcal{N}(0,\Sigma_{\mathrm{true}})$ with an exponential covariance on a one-dimensional index set, $(\Sigma_{\mathrm{true}})_{ij}=\exp(-|i-j|/\tau)$ (unit marginal variance and $\tau=1$). We draw $N=10{,}000$ forecast cases and, for each case, generate an ensemble of size $M=20$ from a misspecified predictive distribution $F_{\mathrm{pred}}=\mathcal{N}(\mu_{\mathrm{pred}},\Sigma_{\mathrm{pred}})$. We consider standard distributional errors that target low-order moments—constant mean shift ($\mu_{\mathrm{pred}}=0.5\mathbf{1}$), variance inflation ($\Sigma_{\mathrm{pred}}$ scaled by $1.75$), and correlation-range misspecification (shorter range, $\tau=0.3$), as well as two PCA-focused covariance perturbations: a \emph{PCA spectrum scramble} that alters the eigenvalue profile of $\Sigma_{\mathrm{true}}$ (while preserving its eigenvectors), and a \emph{PCA-structure} misspecification that preserves marginal variances but rotates principal directions. We evaluate prerank-based projected PIT histograms across these cases, illustrating how different preranks selectively detect distinct types of multivariate misspecification beyond marginal calibration.

\begin{figure}[H]
    \centering
    \includegraphics[width=1\linewidth]{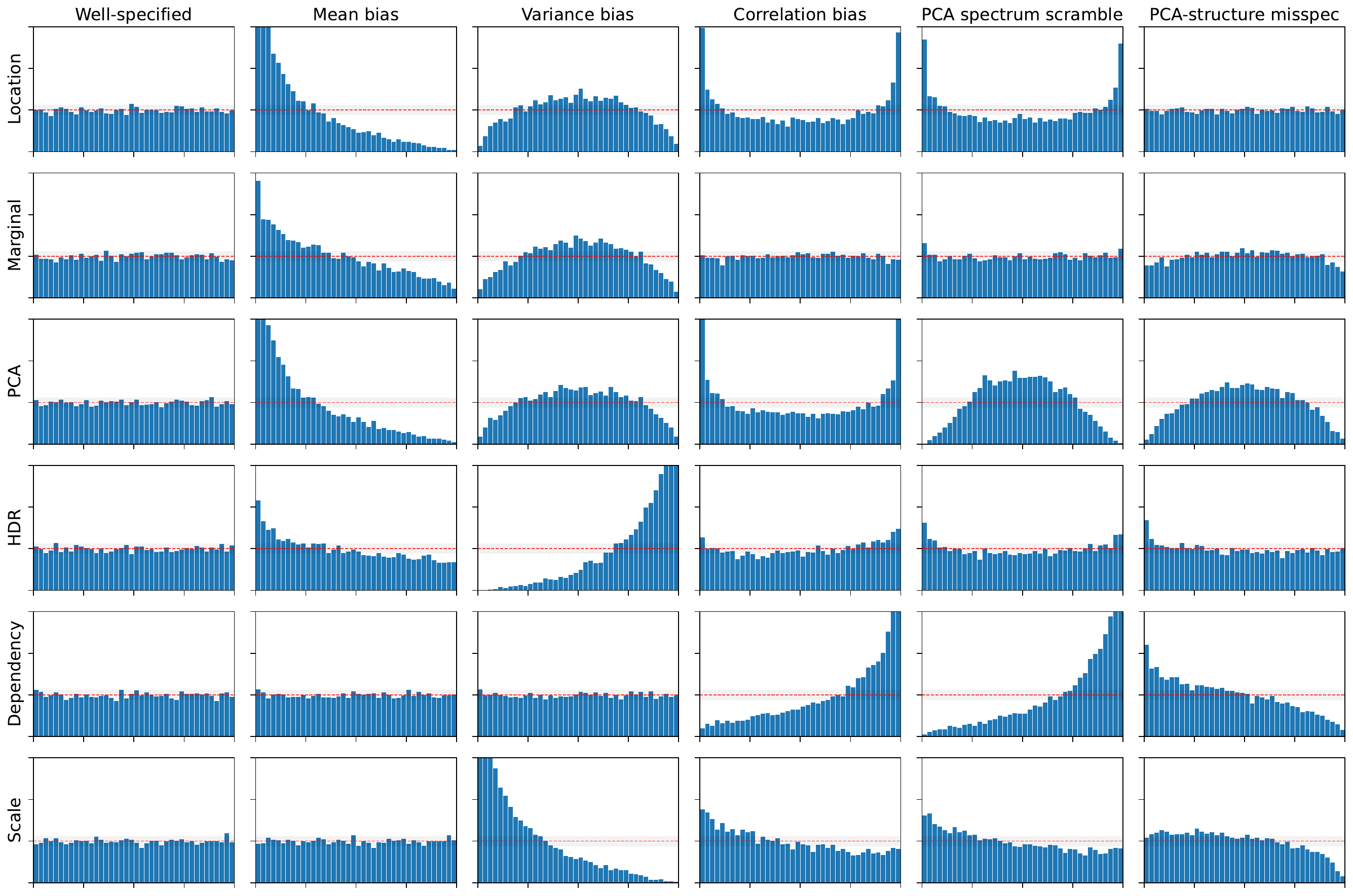}
    \caption{Projected PIT histograms for different pre-rank functions under controlled misspecifications of a Multivariate Gaussian. Each column corresponds to a misspecification scenario. Each row corresponds to a pre-rank function. While most pre-ranks detect marginal and variance-related misspecifications, only the \textbf{PCA pre-rank} consistently reveals miscalibration in scenarios where the covariance structure is altered without affecting marginal distributions.}
    \label{fig:Multivariate_Gaussian}
\end{figure}

\begin{table}[t]
\centering
\caption{Mis-specifications considered in Simulation~1 (Multivariate Gaussian).
$\lambda^{\pi}$ denotes a trace-preserving interpolation between the eigenvalue spectrum of $\Sigma$ and its reversed ordering, and $Q$ is an orthogonal matrix that preserves variance along the mean direction.}
\label{tab:sim1_misspec}
\begin{tabular}{l|ccc}
\toprule
Mis-specification & $\mu$ & $\sigma^2$ & $\tau$ \\
\hline
Mean bias            & $\pm 0.5$ & $1$ & $1$ \\
Variance bias             & $0$ & $\pm 0.75$ & $1$ \\
Correlation bias          & $0$ & $1$ & $\pm 0.7$ \\
PCA-spectrum                       & $0$ & $\lambda^{\pi}$ & $1$ \\
PCA-structure                      & $0$ & $Q\Sigma Q^\top$ & $1$ \\
\bottomrule
\end{tabular}
\end{table}

\paragraph{Simulation 2 (Gaussian random fields).}
We consider a spatially structured $d=25$ dimensional setting by sampling $N=10{,}000$ i.i.d.\ realizations of a Gaussian random field on a $5\times 5$ grid, $y \sim \mathcal{N}(0,\Sigma_{\mathrm{true}})$, where $\Sigma_{\mathrm{true}}$ is an isotropic exponential covariance built from Euclidean distances on the grid (unit marginal variance and range parameter $\tau=1$). To probe which aspects of misspecification are detected by different preranks, we evaluate a suite of predictive models $F_{\mathrm{pred}}=\mathcal{N}(\mu_{\mathrm{pred}},\Sigma_{\mathrm{pred}})$: (i) well-specified, (ii) constant mean shift ($0.5$), (iii) variance inflation ($\times 1.75$), (iv) correlation-range misspecification ($\tau=0.3$), (v) geometric isotropy misspecification (rescaling one spatial axis by a factor $5$), and two PCA-targeted covariance perturbations, PC anisotropy flip and a \emph{PCA-structure} misspecification that preserves marginal variances while altering principal directions. For each case we report prerank-based projected PIT histograms (for Location, Marginal, PCA, Dependency, Scale and HDR), which highlights the selective sensitivity of each prerank to the corresponding type of misspecification in a structured high-dimensional field.

\begin{figure}[h]
    \centering
    \includegraphics[width=1\linewidth]{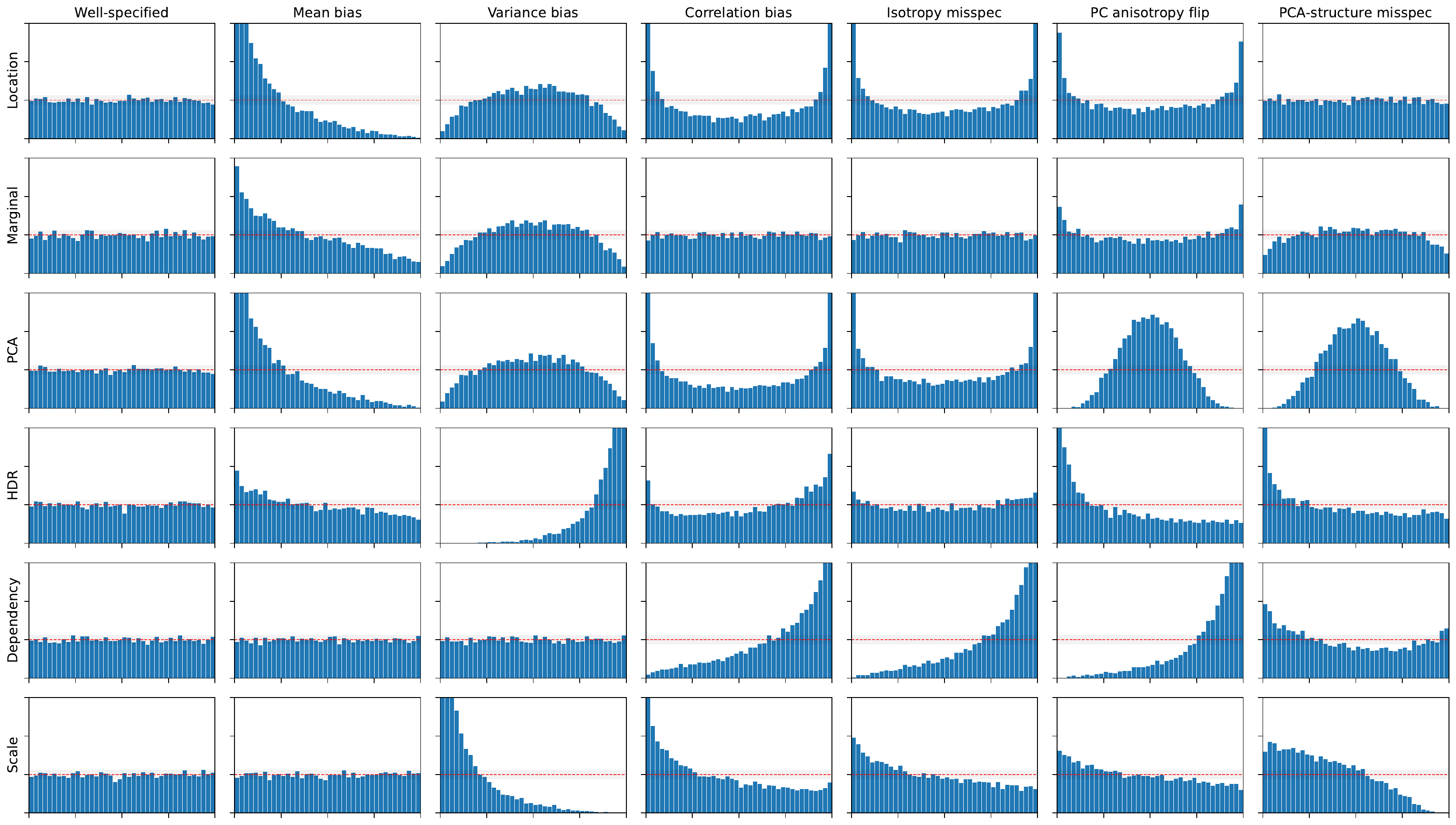}
    \caption{Projected PIT histograms for different pre-rank functions under controlled misspecifications of an example of Gaussian random fields. Each column corresponds to a misspecification scenario. Each row corresponds to a pre-rank function. While most pre-ranks detect marginal and variance-related misspecifications, only the \textbf{PCA pre-rank} consistently reveals miscalibration in scenarios where the covariance structure is altered without affecting marginal distributions.}
    \label{fig:Gaussian_Fields}
\end{figure}

Location pre-rank respond to most misspecifications that affect the mean of the predictive distribution, but again fail to detect the PCA-structure misspecification. Marginal pre-rank detect biases in the mean and  variance, but remain insensitive to correlation, anisotropy, and PCA-based misspecifications. PCA pre-rank consistently detect all misspecifications considered, including errors in spatial dependence, anisotropy, and principal component structure. HDR capture most deviations from the true distribution, but exhibit reduced sensitivity to isotropy misspecification. Dependency pre-rank again display the same previous behavior, but reliably detect errors in spatial correlation, anisotropy, and PCA misspecifications. Finally, scale pre-rank is insensitive to mean bias and, while trivially sensitive to variance misspecification, also respond to spatial correlation, anisotropy, and PCA biases.

\subsection{Additional Misspecifications}
\label{app:cov-misspec}
Let $\Sigma\in\mathbb{R}^{D\times D}$ be a symmetric positive definite covariance matrix.

\paragraph{PCA Spectrum Scramble.} Let the eigendecomposition of $\Sigma$ be :  $\quad \Sigma = U \Lambda U^\top,
\Lambda = \mathrm{diag}(\lambda_1,\dots,\lambda_D),
\quad
\lambda_1 \ge \cdots \ge \lambda_D > 0.$ Define the reversed spectrum $\Lambda^{\mathrm{rev}} := \mathrm{diag}(\lambda_D,\dots,\lambda_1),$ and for a scrambling parameter $\gamma\in[0,1]$, define $\tilde\Lambda(\gamma)
:= (1-\gamma)\Lambda + \gamma \Lambda^{\mathrm{rev}}.$ To preserve total variance, renormalize : $\tilde\Lambda(\gamma)
\;\leftarrow\;
\tilde\Lambda(\gamma)\cdot
\frac{\mathrm{tr}(\Lambda)}{\mathrm{tr}(\tilde\Lambda(\gamma))}.$ The resulting \emph{PCA spectrum--scrambled covariance} is $\Sigma_{\mathrm{scr}}(\gamma)
:=
U\,\tilde\Lambda(\gamma)\,U^\top.$

\paragraph{PCA-Structure Misspecification.} 

Let $e := \frac{1}{\sqrt{D}}(1,\dots,1)^\top$, denote the mean direction, and let $V = [e, v_2,\dots,v_D] \in \mathbb{R}^{D\times D}$, be any orthonormal basis with first column $e$. 

Express $\Sigma$ in this basis: $S := V^\top \Sigma V
 =
\begin{pmatrix}
s_{11} & s_{12}^\top \\
s_{12} & S_\perp
\end{pmatrix},
\qquad
S_\perp \in \mathbb{R}^{(D-1)\times(D-1)}.$ Let the eigendecomposition of $S_\perp$ be $S_\perp
=
W\,\mathrm{diag}(\mu_1,\dots,\mu_{D-1})\,W^\top,
\qquad
\mu_1 \ge \cdots \ge \mu_{D-1}.$ For parameters $c>1$ and $k\le D-1$, define distorted eigenvalues $\tilde\mu_i =
\begin{cases}
c\,\mu_i, & i=1,\dots,k,\\[2pt]
\mu_i/c, & i=D-k,\dots,D-1,\\[2pt]
\mu_i, & \text{otherwise}.
\end{cases}$

Define : $\tilde S_\perp :=
W\,\mathrm{diag}(\tilde\mu_1,\dots,\tilde\mu_{D-1})\,W^\top,
\qquad
\tilde S :=
\begin{pmatrix}
s_{11} & s_{12}^\top \\
s_{12} & \tilde S_\perp
\end{pmatrix}.$

The resulting \emph{PCA-structure-misspecified covariance} is $\Sigma_{\mathrm{pca}}
:=
V\,\tilde S\,V^\top.$ By construction, $e^\top \Sigma_{\mathrm{pca}} e = e^\top \Sigma e,$ while the covariance structure in the subspace orthogonal to $e$ is distorted.

\paragraph{Isotropy Misspecification.}
Let $\{s_i=(x_i,y_i)\}_{i=1}^D \subset \mathbb{R}^2$ denote spatial grid locations and let
\[
\Sigma_{ij}
=
\sigma^2 \exp\!\left(-\frac{\|s_i - s_j\|_2}{\tau}\right)
\]
be the isotropic exponential covariance.
To introduce an example of a geometric isotropy misspecification, we rescale one spatial axis before computing distances.
For a scaling factor $\alpha>1$, define the transformed coordinates
\[
\tilde s_i := (x_i,\;\alpha y_i),
\]
and the corresponding covariance
\[
\Sigma^{\mathrm{aniso}}_{ij}
:=
\sigma^2
\exp\!\left(
-\frac{\|\tilde s_i - \tilde s_j\|_2}{\tau}
\right).
\]
This construction preserves marginal variances while inducing direction-dependent correlation decay.
Such misspecification is invisible to purely marginal diagnostics but alters the spatial dependence
structure and principal directions of variation.

\paragraph{PC Anisotropy Flip.}
Let $\Sigma\in\mathbb{R}^{D\times D}$ be symmetric positive definite with eigendecomposition
\[
\Sigma = U \Lambda U^\top,
\qquad
\Lambda=\mathrm{diag}(\lambda_1,\dots,\lambda_D),
\quad
\lambda_1 \ge \cdots \ge \lambda_D > 0.
\]
We define an example of covariance misspecification that assigns strong variance to the \emph{wrong}
principal directions while preserving total variance.

First, reverse the eigenvalue ordering,
\[
\Lambda^{\mathrm{rev}} := \mathrm{diag}(\lambda_D,\dots,\lambda_1),
\]
and introduce anisotropy by amplifying the leading and shrinking the trailing components.
For an axis ratio $a>1$ and integers $k\le D/2$, define
\[
\tilde\lambda_i =
\begin{cases}
a\,\lambda_{D-i+1}, & i=1,\dots,k,\\[2pt]
\lambda_{D-i+1}, & k<i<D-k,\\[2pt]
\lambda_{D-i+1}/a, & i=D-k+1,\dots,D.
\end{cases}
\]
To preserve total variance, renormalize
\[
\tilde\Lambda
\;\leftarrow\;
\tilde\Lambda \cdot
\frac{\mathrm{tr}(\Lambda)}{\mathrm{tr}(\tilde\Lambda)}.
\]

Finally, to further misalign principal directions, we apply an orthogonal rotation $R$
acting in the $(1,2)$ principal subspace (e.g.\ a $\pi/2$ rotation),
and define
\[
\Sigma_{\mathrm{flip}}
:=
(U R)\,\tilde\Lambda\,(U R)^\top.
\]
This \emph{PC anisotropy flip} preserves marginal variances and overall energy but assigns
large variance to orthogonal directions relative to the true dominant modes.
It is therefore detectable by PCA-based preranks, while potentially evading
mean- or marginal-based calibration diagnostics.


\section{Empirical Experiments}
\label{app:model_hyper}
\paragraph{Neural probabilistic regression model.} \label{mod:model} Our base probabilistic predictor models a conditional predictive distribution as a mixture of $K$ multivariate Gaussian components, where all parameters are generated by a hypernetwork. For each input $x \in \mathcal{X}$ and each mixture component $k \in [K]$, the network predicts the mixture weight $\pi_k(x)$, the mean vector $\mu_k(x) \in \mathbb{R}^D$, and the lower triangular Cholesky factor $L_k(x)$. The covariance matrix is then computed as $\Sigma_k(x) = L_k(x) L_k(x)^\top$, ensuring positive semi-definiteness by construction. The resulting conditional density takes the form:  $\hat{f}_{Y|X=x} = \sum_{k=1}^K \pi_k(x) \, \mathcal{N}(\cdot \mid \mu_k(x), \Sigma_k(x)))$

 
 where $\pi_k(x) \geq 0$ and $\sum_{k=1}^K \pi_k(x) = 1$. We train this model using the NLL scoring rule. 

\paragraph{Hyperparameters.}\label{app:hyperparameters} For the MIX-NLL baseline, we use a mixture of $K = 5$ multivariate Gaussian components. The neural network consists of three fully connected layers with 100 hidden units each, ReLU activations, and is trained using the Adam optimizer with a learning rate of $10^{-4}$. To compute the PCE-KDE regularizer, we estimate the projected PITs $\hat{F}_{T|X}(T)$ using $S = 100$ samples drawn from the predictive distribution with the parameters set to $p=1$ and $M=100$. The temperature parameter $\tau$ in the smoothed indicator function is set to $100$, following prior work in \citet{DheurBenTaieb2023}. 

The regularization strength $\lambda$ in~\eqref{eq:training-objective} controls the degree of calibration enforcement with respect to the chosen pre-rank. As observed in prior work \citep{karandikar2021, wessel2025enforcing}, increasing $\lambda$ typically improves calibration (lower PCE) but may degrade predictive performance (higher NLL or ES). Following the tuning strategy used in \citet{karandikar2021} and \citet{DheurBenTaieb2023}, we select $\lambda$ to minimize PCE while ensuring that ES does not increase by more than 10\% relative to the best ES obtained when $\lambda = 0$. This strategy allows us to improve calibration without sacrificing predictive accuracy. We performed this tuning separately for each dataset and pre-rank pair. We select $\lambda$ on validation set from $\{0, 0.01, 0.1, 1, 5, 10\}$. Table~\ref{tab:lambda-tuning} shows the selected $\lambda$ for each dataset-pre-rank pair. Notably, the majority of selected values are large, often $\lambda = 10$, suggesting that future work could explore larger values or employ more sophisticated tuning strategies such as Bayesian Optimization. In our experiments, we used $\lambda = 10$.

\begin{table}[t]
\centering
\small
\begin{tabular}{l|*{7}{c}}
\hline
Datasets & Marginal & Loc. & Scale & Dep. & PCA & HDR & Copula \\
\hline
households & 10.0 & 10.0 & 10.0 & 5.0 & 10.0 & 10.0 & 5.0\\
air & 10.0 & 10.0 & 10.0 & 10.0 & 10.0 & 10.0 & 5.0\\
births1 & 10.0 & 10.0 & 10.0 & 10.0 & 10.0 & 5.0 & 10.0 \\
births2 & 10.0 & 10.0 & 5.0 & 5.0 & 10.0 & 0.01 & 10.0\\
wage & 10.0 & 10.0 & 5.0 & 10.0 & 5.0 & 1.0 & 10.0\\
scm20d & 10.0 & 10.0 & 10.0 & 10.0 & 5.0 & 10.0 & 0.01\\
scm1d & 10.0 & 10.0 & 10.0 & 10.0 & 10.0 & 0.1 & 10.0 \\
wq & 5.0 & 10.0 & 10.0 & 5.0 & 5.0 & 5.0 & 10.0 \\
scpf & 5.0 & 10.0 & 10.0 & 0.0 & 5.0 & 1.0 & 10.0 \\
meps21 & 5.0 & 5.0 & 5.0 & 10.0 & 10.0& 10.0 & 5.0 \\
meps19 & 5.0 & 10.0 & 1.0 & 10.0 & 10.0 & 10.0 & 10.0\\
meps20 & 5.0 & 10.0 & 1.0 & 1.0 & 10.0 & 10.0 & 10.0\\
house & 5.0 & 10.0 & 5.0 & 10.0 & 5.0 & 5.0 & 10.0 \\
bio & 5.0 & 10.0 & 10.0 & 10.0 & 5.0 & 10.0 & 5.0\\
blog data & 10.0 & 10.0 & 10.0 & 10.0 & 10.0 & 10.0 & 10.0 \\
calcofi & 10.0 & 5.0 & 10.0 & 10.0 & 10.0 & 10.0 & 5.0\\
ansur2 & 10.0 & 5.0 & 5.0 & 10.0 & 10.0 & 5.0 & 10.0\\
taxi & 10.0 & 10.0 & 10.0 & 5.0 & 10.0 & 5.0 & 10.0 \\
\hline
\end{tabular}
\caption{Values of $\lambda$ after hyperparameter tuning with each regularization and each pre-rank. The baseline used is MIX-NLL (Model trained without regularization).}
\label{tab:lambda-tuning}
\end{table}

\paragraph{Distribution of the Test Statistic.}
To assess the statistical significance of observed calibration errors, we characterize the finite-sample distribution of the average Probability Calibration Error (PCE) under the null hypothesis of perfect calibration. For a fixed dataset and pre-rank, the test statistic is defined as the average PCE over the test set, further averaged across five independent training runs.

\begin{figure*}[t]
    \centering
    \includegraphics[width=\textwidth]{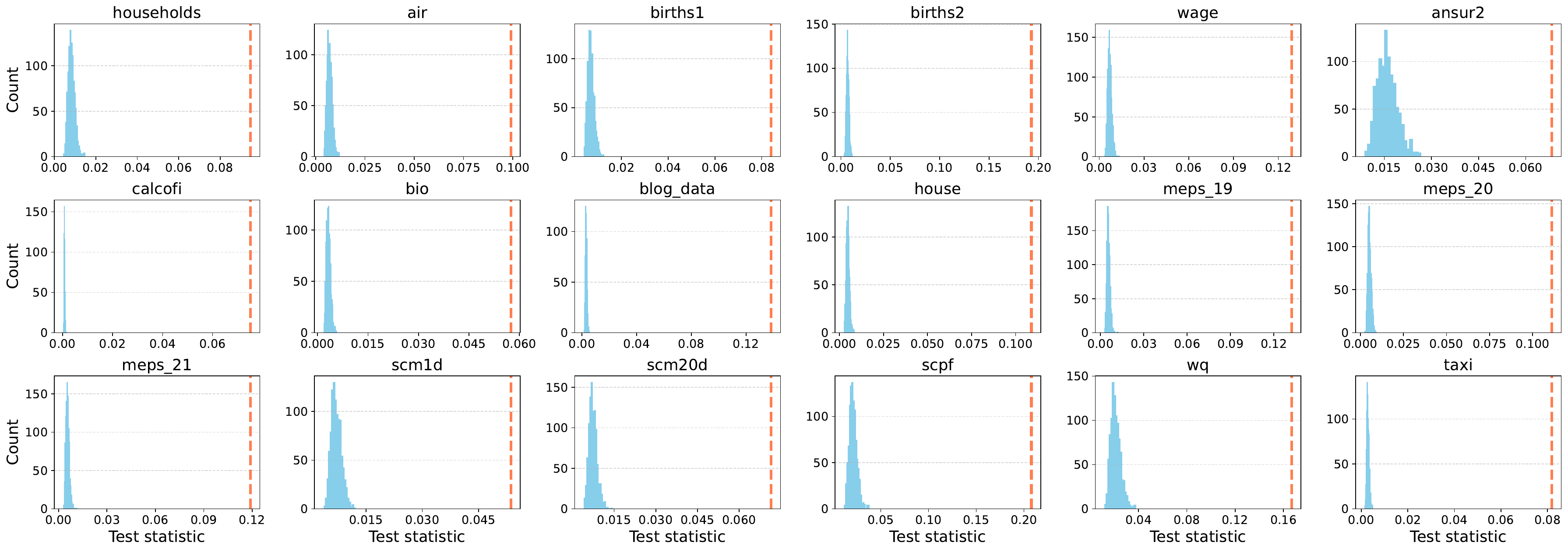}
    \caption{Null distribution (blue histogram) of the average PCE under perfect calibration (i.i.d.\ PIT values drawn from $\mathrm{Unif}(0,1)$ with sample size matching the test set), for each dataset using the \textbf{marginal} pre-rank. The red dashed vertical line marks the observed average PCE on real data (MIX NLL), averaged over five runs.}
    \label{fig:pce_null_vs_real_marginal}
\end{figure*}

\begin{figure*}[t]
    \centering
    \includegraphics[width=\textwidth]{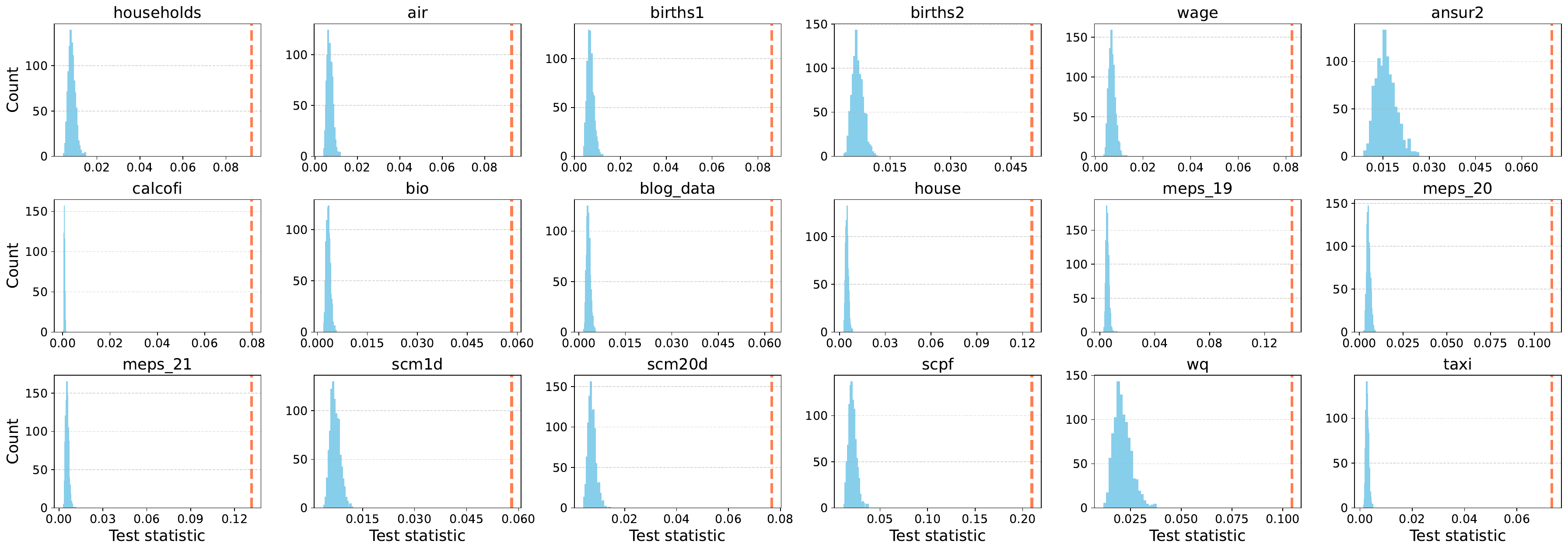}
    \caption{Same as Figure~\ref{fig:pce_null_vs_real_marginal}, but using the \textbf{mean} pre-rank.}
    \label{fig:pce_null_vs_real_mean}
\end{figure*}

\begin{figure*}[t]
    \centering
    \includegraphics[width=\textwidth]{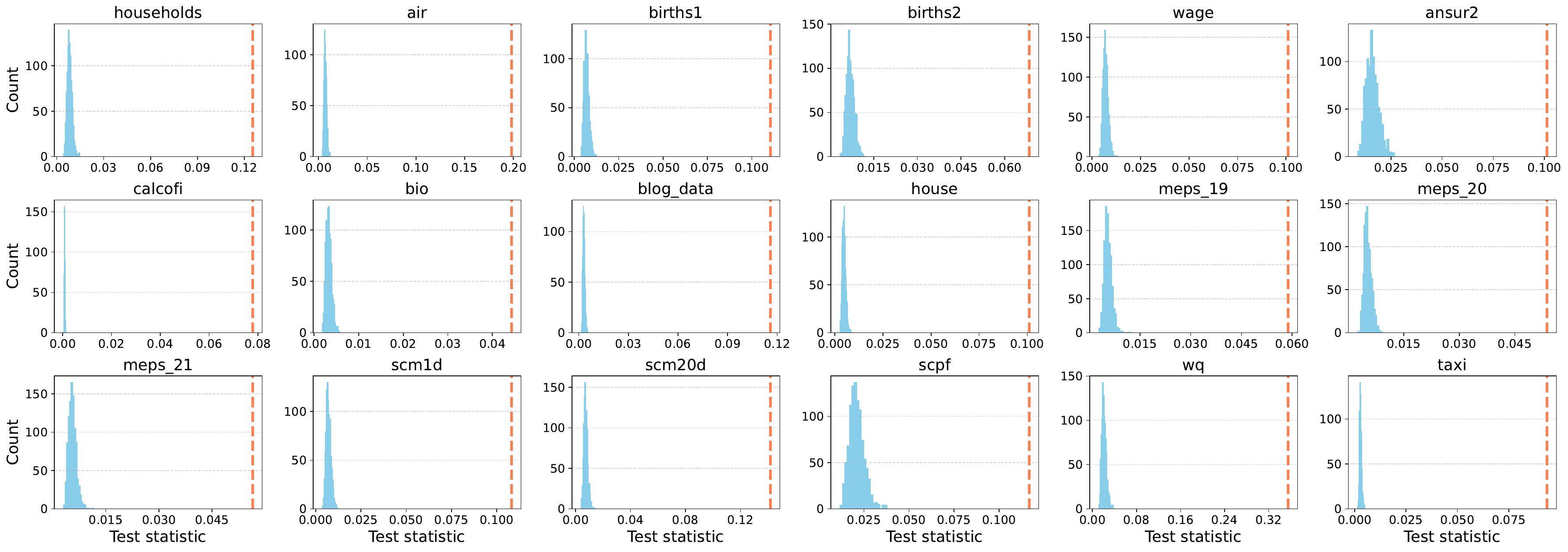}
    \caption{Same as Figure~\ref{fig:pce_null_vs_real_marginal}, but using the \textbf{variance} pre-rank.}
    \label{fig:pce_null_vs_real_variance}
\end{figure*}

\begin{figure*}[t]
    \centering
    \includegraphics[width=\textwidth]{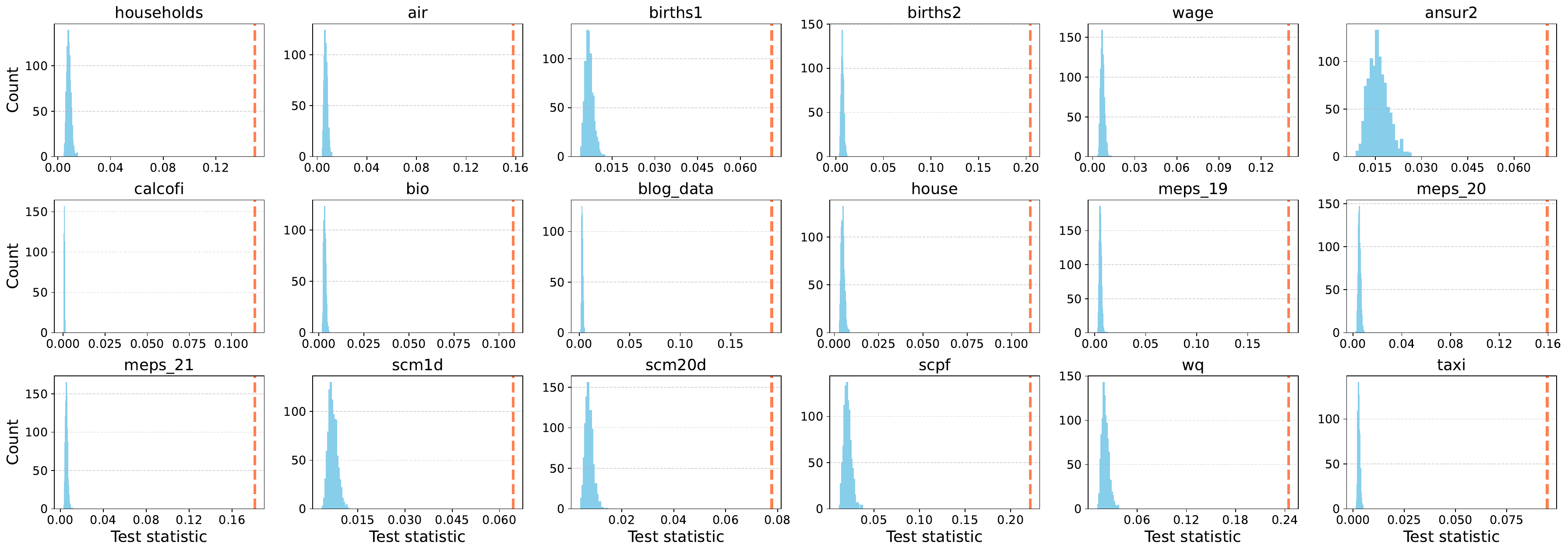}
    \caption{Same as Figure~\ref{fig:pce_null_vs_real_marginal}, but using the \textbf{CDF} pre-rank.}
    \label{fig:pce_null_vs_real_cdf}
\end{figure*}

\begin{figure*}[t]
    \centering
    \includegraphics[width=\textwidth]{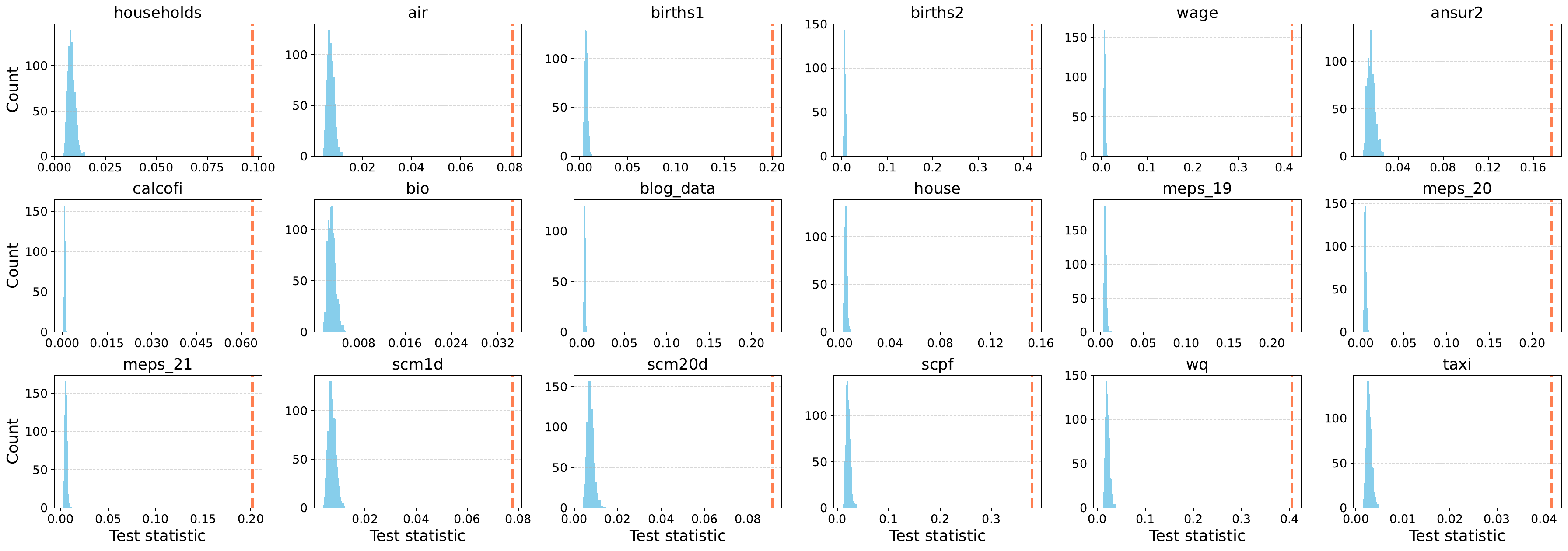}
    \caption{Same as Figure~\ref{fig:pce_null_vs_real_marginal}, but using the \textbf{density} pre-rank.}
    \label{fig:pce_null_vs_real_density}
\end{figure*}

\begin{figure*}[t]
    \centering
    \includegraphics[width=\textwidth]{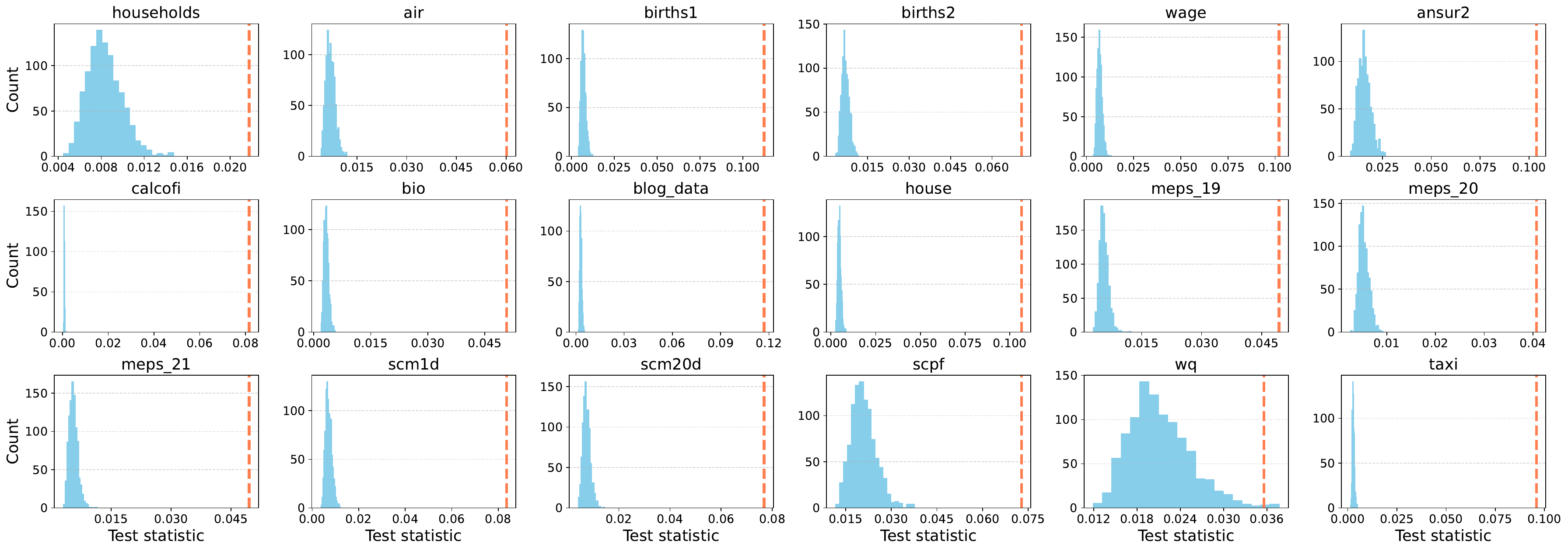}
    \caption{Same as Figure~\ref{fig:pce_null_vs_real_marginal}, but using the \textbf{dependency} pre-rank.}
    \label{fig:pce_null_vs_real_dependency}
\end{figure*}

\begin{figure*}[t]
    \centering
    \includegraphics[width=\textwidth]{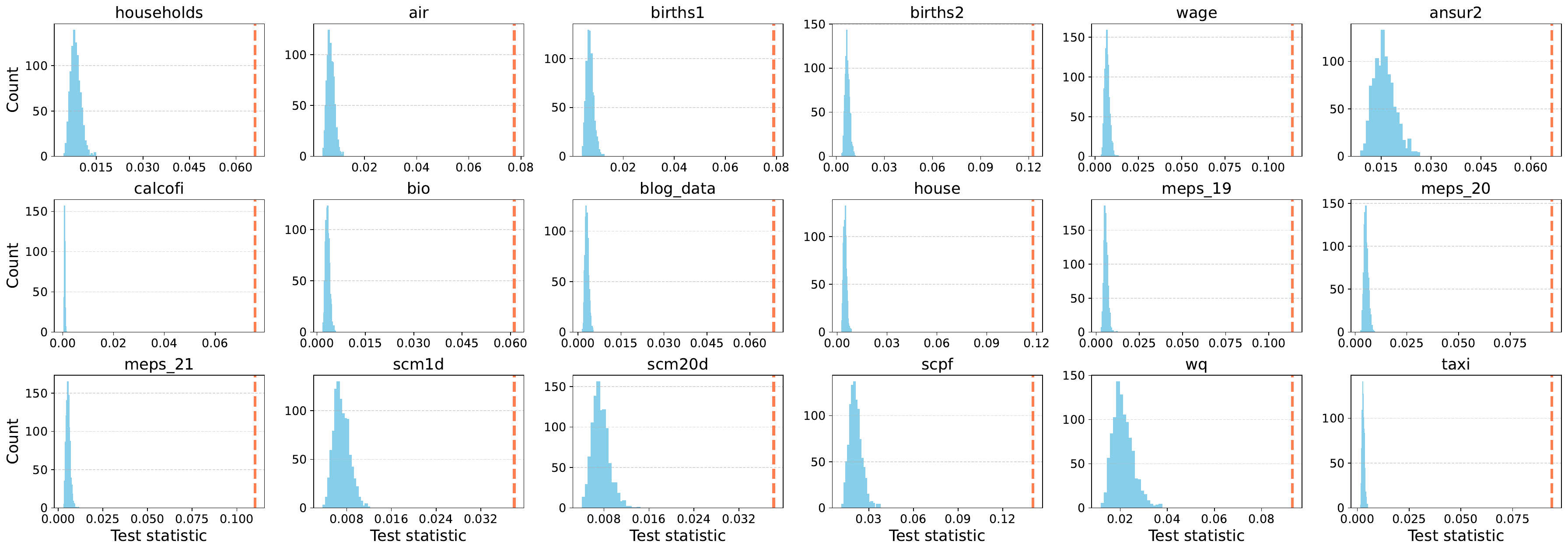}
    \caption{Same as Figure~\ref{fig:pce_null_vs_real_marginal}, but using the \textbf{PCA} pre-rank.}
    \label{fig:pce_null_vs_real_pca}
\end{figure*}

We fix a dataset $d$ with test set size $n_d$ and a given pre-rank $r$. Let $ U^{(k)}_{d,r,1}, \ldots, U^{(k)}_{d,r,n_d}$

denote the Probability Integral Transform (PIT) values obtained from real test observations under pre-rank $r$ in run $k \in \{1,\ldots,5\}$. The empirical test statistic is defined as
\[
\widehat{T}_{d,r}
\;=\;
\frac{1}{5} \sum_{k=1}^{5}
\left(
\frac{1}{n_d} \sum_{i=1}^{n_d}
\mathrm{PCE}\!\left(U^{(k)}_{d,r,i}\right)
\right),
\]
and corresponds to the red dashed vertical line shown in
Figures~\ref{fig:pce_null_vs_real_marginal}--\ref{fig:pce_null_vs_real_pca}.

To characterize the distribution of this statistic under the null hypothesis of perfect calibration, we consider i.i.d.\ random variables $U^{(b)}_{1}, \ldots, U^{(b)}_{n_d}
\;\sim\;
\mathrm{Unif}(0,1),
\qquad b = 1,\ldots,B,$ with $B = 5 \times 10^{4}$. For each Monte Carlo replicate $b$, we compute
\[
T^{(b)}_{d,r}
\;=\;
\frac{1}{n_d} \sum_{i=1}^{n_d}
\mathrm{PCE}\!\left(U^{(b)}_{i}\right),
\]
yielding an empirical approximation of the null distribution $\mathcal{L}_0(T_{d,r})
\;\approx\;
\{T^{(b)}_{d,r}\}_{b=1}^{B}.$

In Figures~\ref{fig:pce_null_vs_real_marginal}--\ref{fig:pce_null_vs_real_pca}, the blue histograms show these null distributions for each dataset and pre-rank, while the red dashed vertical line indicates the observed average PCE computed from real data using the MIX NLL model. The position of the red line relative to the blue distribution provides a direct visual assessment of calibration: values lying in the right tail indicate statistically significant deviations from perfect calibration, while overlap with the bulk of the null distribution indicates limited power to detect miscalibration.




\paragraph{Empirical Calculation of Energy Score.}
We use Energy Score (ES) as a scoring rule metric to evaluate our model performance. ES generalizes Continuous Ranked Probability Score (CRPS) to multivariate settings and is computed empirically as:
\begin{equation}
    \text{ES}(\hat{F}, y) = \frac{1}{G} \sum_{i=1}^G \|\hat{Y}_i - y\| - \frac{1}{2G^2} \sum_{i=1}^G \sum_{j=1}^G \|\hat{Y}_i - \hat{Y}_j\|
    \label{eq:energy}
\end{equation}
where $\{\hat{Y}_i\}_{i=1}^G \sim \hat{F}_{Y|X}$ are $G$ samples drawn from the predictive distribution. We set $G=100$ in all experiments.

\begin{figure}
    \centering
    \includegraphics[width=0.5\linewidth]{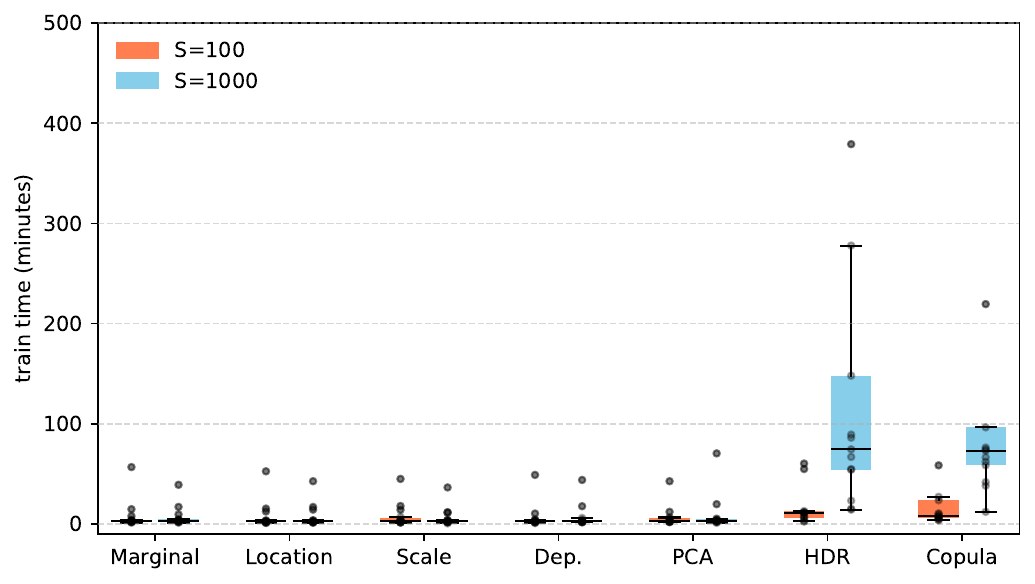}
    \caption{Training time (in minutes) across pre-rank functions for $s=100$ and $s=1000$, aggregated over all datasets. HDR and Copula pre-ranks incur higher computational cost, with PCA remaining substantially more efficient than HDR and Copula pre-rank.}
    \label{fig:sensitivity_s}
\end{figure}


\subsection{Detailed results}
\label{app:results}

\begin{figure}
    \centering
    \includegraphics[width=1\linewidth]{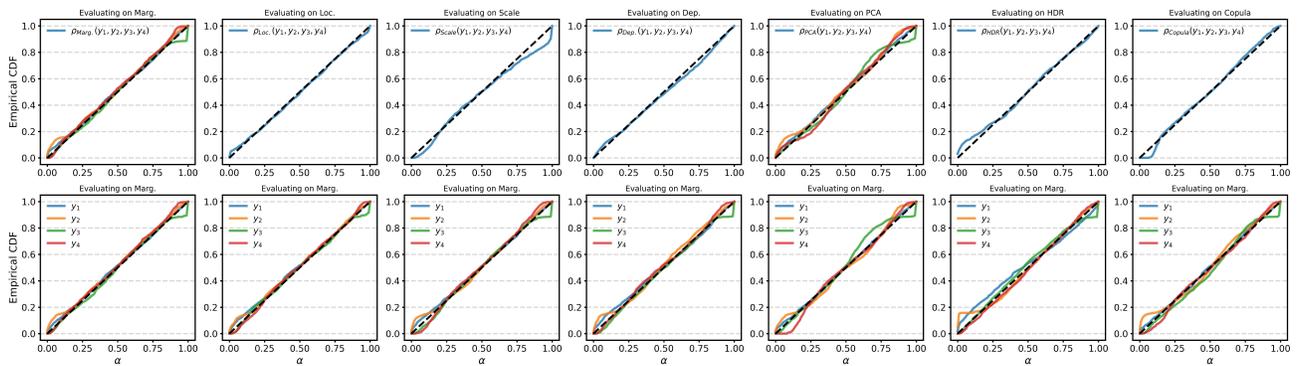}
    \caption{Reliability plots for \textbf{births2} dataset}
    \label{fig:births2}
\end{figure}

\begin{figure}
    \centering
    \includegraphics[width=1\linewidth]{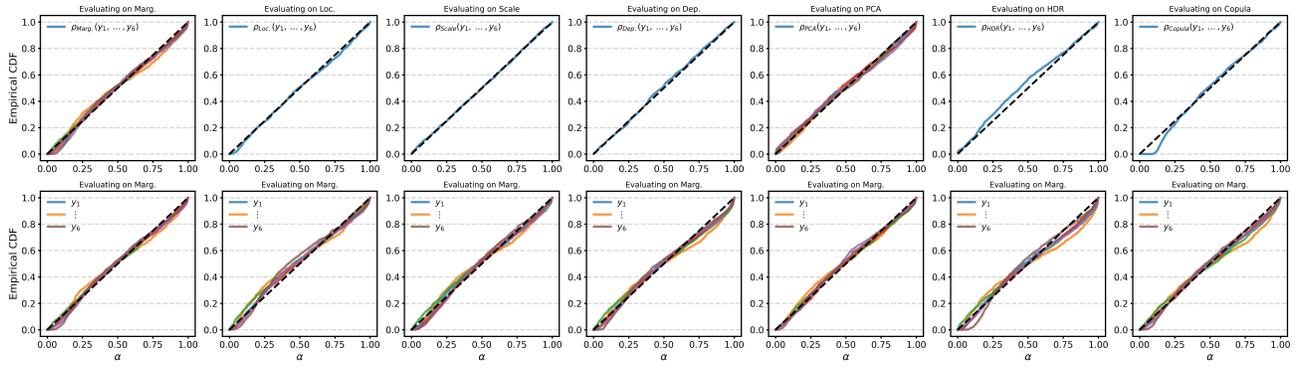}
    \caption{Reliability plots for \textbf{air} dataset}
    \label{fig:air}
\end{figure}

\begin{figure}
    \centering
    \includegraphics[width=1\linewidth]{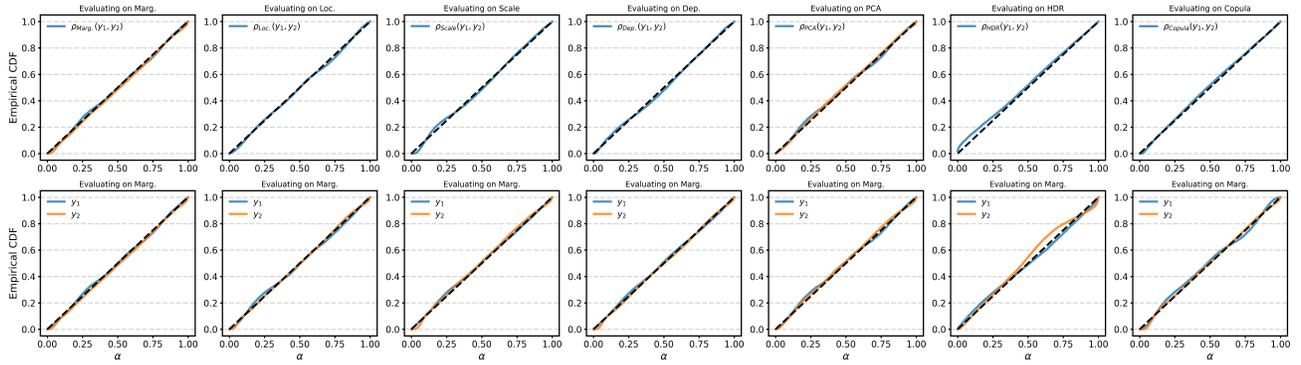}
    \caption{Reliability plots for \textbf{blogdata} dataset}
    \label{fig:blog_data}
\end{figure}

\begin{figure}
    \centering
    \includegraphics[width=1\linewidth]{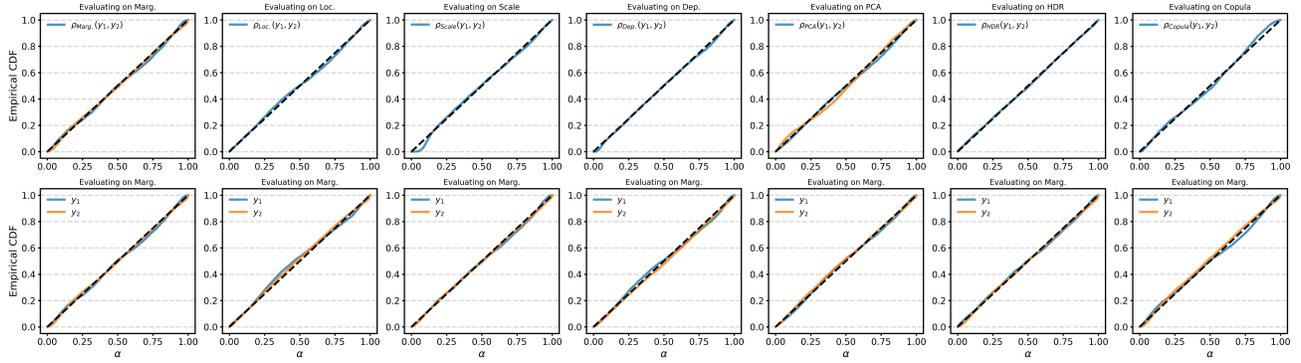}
    \caption{Reliability plots for \textbf{house} dataset}
    \label{fig:house}
\end{figure}

\begin{figure}
    \centering
    \includegraphics[width=1\linewidth]{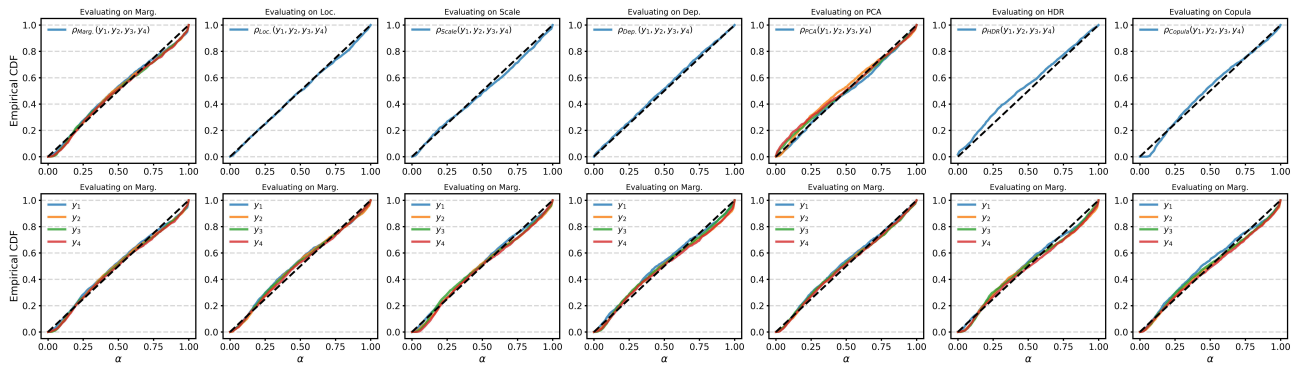}
    \caption{Reliability plots for \textbf{households} dataset}
    \label{fig:households}
\end{figure}

\paragraph{Reliability plots.} Figures~\ref{fig:births2}-\ref{fig:households} show reliability plots obtained by evaluating calibration under different pre-rank functions.
Each column evaluates calibration using a different pre-rank (marginal, location, scale, dependence, PCA-based, HDR-based, and copula-based). The dashed diagonal indicates perfect calibration. Deviations from the diagonal highlight specific forms of miscalibration, illustrating how different pre-ranks emphasize distinct aspects of the joint predictive distribution and reveal complementary calibration behavior.

\paragraph{Pre-rank Calibration.} Table \ref{tab:real-pce-after} reports the exact PCE values averaged over five runs for each pre-rank on which the \texttt{MIX-NLL+PCE-KDE} model was trained using the optimal $\lambda$. For comparison, we also include the PCE values computed with respect to each pre-rank for the baseline \texttt{MIX-NLL} model trained without regularization (see Table \ref{tab:pce_before_formatted}). Note that although the baseline model was not trained with respect to any specific pre-rank, we still evaluate its performance on each pre-rank to highlight the benefit of regularization.



\begin{table}[t]
\centering
\resizebox{\columnwidth}{!}{%
\begin{tabular}{lccccccc}
\toprule
Dataset & Marg. & Loc. & Scale & Dep. & PCA & HDR & Copula \\
\midrule
households & $0.095\,(0.004)$ & $0.092\,(0.002)$ & $0.125\,(0.006)$ & $0.022\,(0.002)$ & $0.066\,(0.002)$ & $0.097\,(0.007)$ & $0.149\,(0.011)$ \\
air        & $0.099\,(0.002)$ & $0.093\,(0.003)$ & $0.198\,(0.005)$ & $0.060\,(0.003)$ & $0.077\,(0.001)$ & $0.081\,(0.007)$ & $0.158\,(0.004)$ \\
births1    & $0.084\,(0.003)$ & $0.086\,(0.003)$ & $0.110\,(0.006)$ & $0.112\,(0.006)$ & $0.079\,(0.002)$ & $0.200\,(0.011)$ & $0.071\,(0.003)$ \\
births2    & $0.193\,(0.008)$ & $0.050\,(0.004)$ & $0.068\,(0.002)$ & $0.071\,(0.002)$ & $0.123\,(0.006)$ & $0.418\,(0.018)$ & $0.204\,(0.031)$ \\
wage       & $0.129\,(0.001)$ & $0.082\,(0.002)$ & $0.101\,(0.004)$ & $0.102\,(0.004)$ & $0.114\,(0.003)$ & $0.417\,(0.005)$ & $0.140\,(0.003)$ \\
scm20d     & $0.072\,(0.001)$ & $0.077\,(0.002)$ & $0.142\,(0.004)$ & $0.077\,(0.007)$ & $0.038\,(0.001)$ & $0.091\,(0.003)$ & $0.078\,(0.001)$ \\
scm1d      & $0.054\,(0.003)$ & $0.058\,(0.005)$ & $0.108\,(0.002)$ & $0.084\,(0.009)$ & $0.038\,(0.001)$ & $0.078\,(0.004)$ & $0.064\,(0.004)$ \\
wq         & $0.167\,(0.005)$ & $0.104\,(0.003)$ & $0.355\,(0.008)$ & $0.036\,(0.001)$ & $0.093\,(0.001)$ & $0.405\,(0.007)$ & $0.245\,(0.001)$ \\
scpf       & $0.208\,(0.009)$ & $0.210\,(0.013)$ & $0.117\,(0.003)$ & $0.073\,(0.006)$ & $0.140\,(0.005)$ & $0.379\,(0.015)$ & $0.222\,(0.019)$ \\
meps21     & $0.119\,(0.002)$ & $0.132\,(0.002)$ & $0.056\,(0.003)$ & $0.050\,(0.004)$ & $0.110\,(0.005)$ & $0.202\,(0.006)$ & $0.181\,(0.006)$ \\
meps19     & $0.132\,(0.004)$ & $0.140\,(0.005)$ & $0.059\,(0.002)$ & $0.049\,(0.002)$ & $0.114\,(0.004)$ & $0.223\,(0.009)$ & $0.190\,(0.006)$ \\
meps20     & $0.111\,(0.004)$ & $0.110\,(0.007)$ & $0.054\,(0.001)$ & $0.041\,(0.001)$ & $0.095\,(0.004)$ & $0.223\,(0.007)$ & $0.159\,(0.005)$ \\
house      & $0.109\,(0.002)$ & $0.126\,(0.002)$ & $0.101\,(0.001)$ & $0.107\,(0.001)$ & $0.118\,(0.002)$ & $0.153\,(0.003)$ & $0.110\,(0.002)$ \\
bio        & $0.057\,(0.002)$ & $0.058\,(0.005)$ & $0.044\,(0.002)$ & $0.051\,(0.002)$ & $0.061\,(0.003)$ & $0.034\,(0.002)$ & $0.108\,(0.005)$ \\
blogdata   & $0.138\,(0.002)$ & $0.062\,(0.002)$ & $0.116\,(0.004)$ & $0.117\,(0.004)$ & $0.068\,(0.001)$ & $0.224\,(0.003)$ & $0.191\,(0.006)$ \\
calcofi    & $0.075\,(0.000)$ & $0.080\,(0.001)$ & $0.078\,(0.000)$ & $0.082\,(0.000)$ & $0.076\,(0.000)$ & $0.064\,(0.001)$ & $0.114\,(0.001)$ \\
ansur2     & $0.068\,(0.004)$ & $0.070\,(0.006)$ & $0.101\,(0.006)$ & $0.104\,(0.006)$ & $0.066\,(0.004)$ & $0.176\,(0.011)$ & $0.071\,(0.005)$ \\
taxi       & $0.082\,(0.001)$ & $0.073\,(0.001)$ & $0.094\,(0.003)$ & $0.096\,(0.003)$ & $0.094\,(0.002)$ & $0.042\,(0.001)$ & $0.095\,(0.002)$ \\
\bottomrule
\end{tabular}
}
\caption{\textbf{PCE} results of real-world experiments using the\textbf{ MIX-NLL} model. PCE values are computed using seven pre-rank functions across 18 real datasets and averaged over five runs. Standard errors are shown in parentheses.}
\label{tab:pce_before_formatted}
\end{table}

 \begin{table}[t]
\centering
\resizebox{\columnwidth}{!}{%
\begin{tabular}{lccccccc}
\toprule
Dataset & Marg. & Loc. & Scale & Dep. & PCA & HDR & Copula \\
\midrule
households & $0.030\,(0.001)$ & $0.024\,(0.002)$ & $0.021\,(0.002)$ & $0.021\,(0.003)$ & $0.026\,(0.001)$ & $0.039\,(0.003)$ & $0.031\,(0.001)$ \\
air        & $0.040\,(0.001)$ & $0.026\,(0.002)$ & $0.027\,(0.001)$ & $0.027\,(0.002)$ & $0.039\,(0.002)$ & $0.045\,(0.005)$ & $0.029\,(0.001)$ \\
births1    & $0.028\,(0.001)$ & $0.027\,(0.002)$ & $0.031\,(0.002)$ & $0.027\,(0.002)$ & $0.034\,(0.000)$ & $0.034\,(0.003)$ & $0.027\,(0.001)$ \\
births2    & $0.031\,(0.002)$ & $0.029\,(0.002)$ & $0.045\,(0.002)$ & $0.032\,(0.001)$ & $0.052\,(0.006)$ & $0.428\,(0.002)$ & $0.033\,(0.002)$ \\
wage       & $0.051\,(0.020)$ & $0.044\,(0.013)$ & $0.025\,(0.001)$ & $0.024\,(0.002)$ & $0.052\,(0.008)$ & $0.364\,(0.012)$ & $0.095\,(0.015)$ \\
scm20d     & $0.033\,(0.002)$ & $0.025\,(0.001)$ & $0.038\,(0.003)$ & $0.022\,(0.002)$ & $0.033\,(0.001)$ & $0.093\,(0.003)$ & $0.029\,(0.001)$ \\
scm1d      & $0.025\,(0.001)$ & $0.024\,(0.002)$ & $0.036\,(0.003)$ & $0.020\,(0.002)$ & $0.035\,(0.001)$ & $0.090\,(0.015)$ & $0.036\,(0.007)$ \\
wq         & $0.154\,(0.007)$ & $0.076\,(0.010)$ & $0.311\,(0.018)$ & $0.028\,(0.003)$ & $0.091\,(0.004)$ & $0.376\,(0.026)$ & $0.244\,(0.001)$ \\
scpf       & $0.039\,(0.002)$ & $0.026\,(0.003)$ & $0.041\,(0.005)$ & $0.081\,(0.005)$ & $0.063\,(0.004)$ & $0.299\,(0.010)$ & $0.032\,(0.006)$ \\
meps21     & $0.026\,(0.001)$ & $0.025\,(0.001)$ & $0.031\,(0.001)$ & $0.024\,(0.001)$ & $0.025\,(0.001)$ & $0.032\,(0.001)$ & $0.023\,(0.001)$ \\
meps19     & $0.026\,(0.002)$ & $0.023\,(0.001)$ & $0.050\,(0.002)$ & $0.025\,(0.002)$ & $0.024\,(0.001)$ & $0.031\,(0.001)$ & $0.022\,(0.001)$ \\
meps20     & $0.024\,(0.001)$ & $0.023\,(0.001)$ & $0.049\,(0.001)$ & $0.033\,(0.001)$ & $0.024\,(0.000)$ & $0.034\,(0.005)$ & $0.024\,(0.001)$ \\
house      & $0.027\,(0.003)$ & $0.020\,(0.001)$ & $0.025\,(0.003)$ & $0.020\,(0.001)$ & $0.033\,(0.004)$ & $0.028\,(0.002)$ & $0.021\,(0.000)$ \\
bio        & $0.021\,(0.001)$ & $0.020\,(0.001)$ & $0.021\,(0.001)$ & $0.021\,(0.001)$ & $0.021\,(0.001)$ & $0.021\,(0.001)$ & $0.021\,(0.001)$ \\
blogdata   & $0.023\,(0.001)$ & $0.024\,(0.000)$ & $0.023\,(0.001)$ & $0.023\,(0.001)$ & $0.025\,(0.000)$ & $0.030\,(0.001)$ & $0.024\,(0.001)$ \\
calcofi    & $0.020\,(0.000)$ & $0.021\,(0.000)$ & $0.020\,(0.000)$ & $0.020\,(0.000)$ & $0.021\,(0.000)$ & $0.020\,(0.000)$ & $0.020\,(0.000)$ \\
ansur2     & $0.032\,(0.004)$ & $0.040\,(0.009)$ & $0.031\,(0.004)$ & $0.025\,(0.003)$ & $0.038\,(0.003)$ & $0.040\,(0.015)$ & $0.039\,(0.009)$ \\
taxi       & $0.021\,(0.001)$ & $0.022\,(0.001)$ & $0.025\,(0.000)$ & $0.023\,(0.001)$ & $0.022\,(0.001)$ & $0.022\,(0.000)$ & $0.022\,(0.001)$ \\
\bottomrule
\end{tabular}
}
\caption{\textbf{PCE} results after applying \textbf{PCE-KDE regularization with MIX-NLL} using the optimal $\lambda$. Results are reported for seven pre-rank functions across 18 real datasets, averaged over five runs. Standard errors are shown in parentheses.}
\label{tab:real-pce-after}
\end{table}

\begin{table}[t]
\centering
\resizebox{\columnwidth}{!}{%
\begin{tabular}{lccccccc}
\toprule
Dataset & Marg. & Loc. & Scale & Dep. & PCA & HDR & Copula \\
\midrule
households & $3.33\,(0.08)$ & $3.40\,(0.07)$ & $3.34\,(0.08)$ & $3.23\,(0.11)$ & $3.33\,(0.15)$ & $3.29\,(0.09)$ & $3.35\,(0.11)$ \\
air        & $5.80\,(0.19)$ & $5.81\,(0.19)$ & $5.75\,(0.20)$ & $5.71\,(0.18)$ & $5.79\,(0.19)$ & $5.67\,(0.16)$ & $5.80\,(0.21)$ \\
births1    & $1.82\,(0.39)$ & $1.37\,(0.19)$ & $1.74\,(0.44)$ & $1.63\,(0.27)$ & $1.36\,(0.21)$ & $2.12\,(0.09)$ & $1.39\,(0.17)$ \\
births2    & $-4.28\,(0.36)$ & $-3.72\,(1.76)$ & $-4.71\,(0.39)$ & $-4.33\,(0.45)$ & $-3.91\,(0.76)$ & $-4.10\,(0.84)$ & $-4.35\,(0.72)$ \\
wage       & $1.63\,(0.96)$ & $1.20\,(0.96)$ & $0.77\,(0.32)$ & $0.81\,(0.15)$ & $1.01\,(0.76)$ & $0.61\,(0.12)$ & $2.55\,(0.71)$ \\
scm20d     & $7.70\,(0.32)$ & $7.89\,(0.18)$ & $7.72\,(0.30)$ & $7.67\,(0.22)$ & $7.56\,(0.21)$ & $7.89\,(0.18)$ & $7.69\,(0.13)$ \\
scm1d      & $7.02\,(1.35)$ & $6.58\,(0.50)$ & $6.77\,(1.31)$ & $7.17\,(0.70)$ & $11.35\,(5.23)$ & $8.04\,(1.63)$ & $10.34\,(5.82)$ \\
wq         & $26.33\,(7.30)$ & $25.08\,(6.20)$ & $27.51\,(8.82)$ & $37.05\,(34.21)$ & $21.94\,(1.62)$ & $30.10\,(12.71)$ & $47.46\,(38.00)$ \\
scpf       & $-3.52\,(2.14)$ & $-3.97\,(1.32)$ & $-4.48\,(0.88)$ & $-4.14\,(1.27)$ & $-4.35\,(0.93)$ & $-4.74\,(1.40)$ & $-3.69\,(2.48)$ \\
meps21     & $-1.79\,(0.18)$ & $-1.79\,(0.06)$ & $-1.74\,(0.10)$ & $-1.72\,(0.11)$ & $-1.68\,(0.05)$ & $-1.06\,(0.21)$ & $-1.83\,(0.09)$ \\
meps19     & $-1.84\,(0.09)$ & $-1.81\,(0.06)$ & $-2.02\,(0.13)$ & $-1.71\,(0.13)$ & $-1.74\,(0.10)$ & $-1.23\,(0.11)$ & $-1.77\,(0.13)$ \\
meps20     & $-1.75\,(0.15)$ & $-1.71\,(0.13)$ & $-1.85\,(0.17)$ & $-1.79\,(0.25)$ & $-1.67\,(0.11)$ & $-1.01\,(0.21)$ & $-1.62\,(0.20)$ \\
house      & $0.97\,(0.46)$ & $0.99\,(0.37)$ & $1.86\,(1.70)$ & $0.66\,(0.07)$ & $0.95\,(0.36)$ & $8.31\,(10.85)$ & $0.98\,(0.33)$ \\
bio        & $-0.69\,(0.06)$ & $-0.64\,(0.12)$ & $-0.66\,(0.08)$ & $-0.60\,(0.17)$ & $-0.67\,(0.08)$ & $-0.48\,(0.14)$ & $-0.67\,(0.11)$ \\
blogdata   & $-0.37\,(0.07)$ & $-0.45\,(0.04)$ & $-0.45\,(0.08)$ & $-0.51\,(0.12)$ & $-0.35\,(0.12)$ & $-1.30\,(0.52)$ & $-0.62\,(0.20)$ \\
calcofi    & $0.59\,(0.01)$ & $0.59\,(0.00)$ & $0.60\,(0.01)$ & $0.60\,(0.01)$ & $0.60\,(0.01)$ & $0.60\,(0.01)$ & $0.59\,(0.00)$ \\
ansur2     & $1.87\,(0.04)$ & $1.86\,(0.05)$ & $1.82\,(0.04)$ & $1.83\,(0.05)$ & $1.87\,(0.05)$ & $1.96\,(0.10)$ & $1.87\,(0.04)$ \\
taxi       & $1.79\,(0.03)$ & $1.79\,(0.03)$ & $1.76\,(0.03)$ & $1.76\,(0.04)$ & $1.81\,(0.02)$ & $1.74\,(0.02)$ & $1.81\,(0.02)$ \\
\bottomrule
\end{tabular}
}

\caption{\textbf{NLL} results after applying \textbf{PCE-KDE regularization with MIX-NLL} using the optimal $\lambda$. Results are reported for seven pre-rank functions across 18 real datasets, averaged over five runs. Standard errors are shown in parentheses.}
\label{tab:real-nll-after}
\end{table}

\begin{table}[t]
\centering
\resizebox{\columnwidth}{!}{%
\begin{tabular}{lccccccc}
\toprule
Dataset & Marg. & Loc. & Scale & Dep. & PCA & HDR & Copula \\
\midrule
households & $0.939\,(0.034)$ & $0.944\,(0.024)$ & $0.937\,(0.028)$ & $0.921\,(0.031)$ & $0.937\,(0.036)$ & $0.933\,(0.037)$ & $0.952\,(0.040)$ \\
air        & $1.421\,(0.041)$ & $1.427\,(0.030)$ & $1.418\,(0.049)$ & $1.406\,(0.041)$ & $1.417\,(0.043)$ & $1.405\,(0.031)$ & $1.435\,(0.042)$ \\
births1    & $0.724\,(0.013)$ & $0.734\,(0.017)$ & $0.718\,(0.012)$ & $0.714\,(0.014)$ & $0.730\,(0.010)$ & $0.712\,(0.007)$ & $0.729\,(0.009)$ \\
births2    & $0.964\,(0.067)$ & $0.940\,(0.060)$ & $0.905\,(0.061)$ & $0.924\,(0.055)$ & $0.930\,(0.064)$ & $0.922\,(0.060)$ & $0.960\,(0.052)$ \\
wage       & $0.762\,(0.082)$ & $0.766\,(0.083)$ & $0.714\,(0.020)$ & $0.738\,(0.018)$ & $0.719\,(0.013)$ & $0.704\,(0.010)$ & $0.878\,(0.079)$ \\
scm20d     & $2.247\,(0.049)$ & $2.295\,(0.042)$ & $2.379\,(0.033)$ & $2.358\,(0.046)$ & $2.297\,(0.066)$ & $2.384\,(0.017)$ & $2.350\,(0.063)$ \\
scm1d      & $1.886\,(0.078)$ & $1.925\,(0.084)$ & $2.060\,(0.097)$ & $2.105\,(0.114)$ & $1.960\,(0.066)$ & $1.993\,(0.152)$ & $2.072\,(0.086)$ \\
wq         & $2.597\,(0.095)$ & $2.592\,(0.098)$ & $2.601\,(0.093)$ & $2.599\,(0.099)$ & $2.599\,(0.090)$ & $2.595\,(0.096)$ & $2.596\,(0.078)$ \\
scpf       & $0.493\,(0.283)$ & $0.497\,(0.271)$ & $0.487\,(0.259)$ & $0.480\,(0.261)$ & $0.492\,(0.262)$ & $0.482\,(0.268)$ & $0.507\,(0.283)$ \\
meps21     & $0.410\,(0.017)$ & $0.402\,(0.004)$ & $0.423\,(0.013)$ & $0.437\,(0.018)$ & $0.415\,(0.011)$ & $0.413\,(0.023)$ & $0.413\,(0.028)$ \\
meps19     & $0.392\,(0.022)$ & $0.401\,(0.027)$ & $0.376\,(0.019)$ & $0.423\,(0.023)$ & $0.405\,(0.017)$ & $0.395\,(0.018)$ & $0.412\,(0.018)$ \\
meps20     & $0.405\,(0.033)$ & $0.421\,(0.028)$ & $0.390\,(0.027)$ & $0.402\,(0.038)$ & $0.428\,(0.032)$ & $0.425\,(0.024)$ & $0.426\,(0.029)$ \\
house      & $0.571\,(0.067)$ & $0.574\,(0.054)$ & $0.576\,(0.068)$ & $0.535\,(0.015)$ & $0.565\,(0.049)$ & $0.574\,(0.037)$ & $0.576\,(0.056)$ \\
bio        & $0.245\,(0.016)$ & $0.248\,(0.016)$ & $0.250\,(0.015)$ & $0.253\,(0.019)$ & $0.246\,(0.014)$ & $0.258\,(0.017)$ & $0.247\,(0.014)$ \\
blogdata   & $0.637\,(0.012)$ & $0.680\,(0.010)$ & $0.707\,(0.020)$ & $0.690\,(0.015)$ & $0.668\,(0.015)$ & $0.376\,(0.131)$ & $0.635\,(0.004)$ \\
calcofi    & $0.421\,(0.001)$ & $0.421\,(0.001)$ & $0.421\,(0.001)$ & $0.421\,(0.001)$ & $0.421\,(0.001)$ & $0.421\,(0.001)$ & $0.422\,(0.001)$ \\
ansur2     & $0.544\,(0.011)$ & $0.546\,(0.017)$ & $0.541\,(0.012)$ & $0.540\,(0.014)$ & $0.549\,(0.018)$ & $0.577\,(0.048)$ & $0.546\,(0.012)$ \\
taxi       & $0.706\,(0.011)$ & $0.703\,(0.008)$ & $0.707\,(0.007)$ & $0.701\,(0.007)$ & $0.707\,(0.007)$ & $0.698\,(0.010)$ & $0.703\,(0.010)$ \\
\bottomrule
\end{tabular}
}
\caption{\textbf{Energy score (ES)} results after applying \textbf{PCE-KDE regularization with MIX-NLL} using the optimal $\lambda$. Results are reported for seven pre-rank functions across 18 real datasets, averaged over five runs. Standard errors are shown in parentheses.}
\label{tab:real-energy-after}
\end{table}

\begin{table}[t]
\centering
\resizebox{\columnwidth}{!}{%
\begin{tabular}{lccccccc}
\toprule
Dataset & Copula & density & Dependency & Marginal & Mean & PCA & Scale \\
\midrule
air       & $0.143\,(0.002)$ & $0.074\,(0.002)$ & $\mathbf{0.050\,(0.001)}$ & $0.091\,(0.002)$ & $0.085\,(0.003)$ & $0.071\,(0.002)$ & $0.191\,(0.003)$ \\
ansur2    & $0.078\,(0.003)$ & $0.160\,(0.002)$ & $0.094\,(0.003)$ & $0.067\,(0.002)$ & $0.071\,(0.002)$ & $\mathbf{0.066\,(0.002)}$ & $0.095\,(0.003)$ \\
bio       & $0.041\,(0.002)$ & $0.032\,(0.001)$ & $0.036\,(0.001)$ & $0.038\,(0.001)$ & $\mathbf{0.021\,(0.001)}$ & $0.031\,(0.001)$ & $0.030\,(0.001)$ \\
births1   & $0.063\,(0.002)$ & $0.202\,(0.003)$ & $0.094\,(0.002)$ & $0.070\,(0.002)$ & $0.071\,(0.002)$ & $\mathbf{0.071\,(0.001)}$ & $0.093\,(0.002)$ \\
births2   & $0.105\,(0.002)$ & $0.397\,(0.003)$ & $0.071\,(0.002)$ & $0.174\,(0.002)$ & $\mathbf{0.041\,(0.002)}$ & $0.108\,(0.003)$ & $0.058\,(0.002)$ \\
blog\_data& $0.037\,(0.002)$ & $0.202\,(0.003)$ & $0.116\,(0.003)$ & $0.077\,(0.002)$ & $0.061\,(0.002)$ & $\mathbf{0.057\,(0.002)}$ & $0.115\,(0.003)$ \\
calcofi   & $0.047\,(0.002)$ & $\mathbf{0.028\,(0.001)}$ & $0.033\,(0.001)$ & $0.038\,(0.001)$ & $0.028\,(0.001)$ & $0.037\,(0.001)$ & $0.031\,(0.001)$ \\
house     & $0.091\,(0.002)$ & $0.119\,(0.003)$ & $0.091\,(0.002)$ & $0.093\,(0.002)$ & $0.113\,(0.002)$ & $0.102\,(0.002)$ & $\mathbf{0.087\,(0.002)}$ \\
households& $0.137\,(0.002)$ & $0.088\,(0.002)$ & $\mathbf{0.020\,(0.001)}$ & $0.089\,(0.002)$ & $0.089\,(0.002)$ & $0.066\,(0.002)$ & $0.123\,(0.002)$ \\
meps\_19  & $0.058\,(0.002)$ & $0.217\,(0.003)$ & $\mathbf{0.040\,(0.002)}$ & $0.060\,(0.002)$ & $0.053\,(0.002)$ & $0.042\,(0.002)$ & $0.055\,(0.002)$ \\
meps\_20  & $0.044\,(0.002)$ & $0.207\,(0.003)$ & $\mathbf{0.033\,(0.002)}$ & $0.054\,(0.002)$ & $0.037\,(0.002)$ & $0.034\,(0.002)$ & $0.051\,(0.002)$ \\
meps\_21  & $0.054\,(0.002)$ & $0.208\,(0.003)$ & $\mathbf{0.038\,(0.002)}$ & $0.056\,(0.002)$ & $0.045\,(0.002)$ & $0.035\,(0.002)$ & $0.054\,(0.003)$ \\
scm1d     & $0.062\,(0.002)$ & $0.077\,(0.002)$ & $0.075\,(0.002)$ & $0.050\,(0.002)$ & $0.053\,(0.002)$ & $\mathbf{0.038\,(0.002)}$ & $0.104\,(0.003)$ \\
scm20d    & $0.072\,(0.002)$ & $0.088\,(0.002)$ & $0.068\,(0.002)$ & $0.068\,(0.002)$ & $0.074\,(0.002)$ & $\mathbf{0.037\,(0.001)}$ & $0.130\,(0.003)$ \\
scpf      & $0.089\,(0.003)$ & $0.343\,(0.003)$ & $\mathbf{0.071\,(0.003)}$ & $0.125\,(0.003)$ & $0.100\,(0.003)$ & $0.106\,(0.003)$ & $0.119\,(0.003)$ \\
taxi      & $0.083\,(0.002)$ & $0.034\,(0.001)$ & $0.081\,(0.002)$ & $0.073\,(0.002)$ & $0.067\,(0.002)$ & $0.078\,(0.002)$ & $\mathbf{0.079\,(0.002)}$ \\
wage      & $0.135\,(0.003)$ & $0.412\,(0.003)$ & $0.093\,(0.003)$ & $0.127\,(0.003)$ & $\mathbf{0.078\,(0.003)}$ & $0.113\,(0.003)$ & $0.092\,(0.003)$ \\
wq        & $0.245\,(0.003)$ & $0.399\,(0.003)$ & $\mathbf{0.026\,(0.001)}$ & $0.167\,(0.003)$ & $0.111\,(0.003)$ & $0.093\,(0.003)$ & $0.350\,(0.003)$ \\
\bottomrule
\end{tabular}
}
\caption{PCEs when the model is trained with \textbf{Marginal+ALL} and evaluated on each simple pre-rank. Each entry shows mean PCE (std) over five runs. Bold indicates the best (lowest) PCE within each dataset row.}
\label{tab:pce_marginalALL_simple_best_3dp}
\end{table}

\textbf{\begin{table}[t]
\centering
\resizebox{\columnwidth}{!}{%
\begin{tabular}{lccccccc}
\toprule
Dataset & Copula & density & Dependency & Marginal & Mean & PCA & Scale \\
\midrule
air       & $0.144\,(0.002)$ & $0.081\,(0.003)$ & $\mathbf{0.048\,(0.001)}$ & $0.093\,(0.002)$ & $0.088\,(0.002)$ & $0.071\,(0.001)$ & $0.190\,(0.003)$ \\
ansur2    & $0.070\,(0.003)$ & $0.152\,(0.002)$ & $0.092\,(0.003)$ & $0.064\,(0.002)$ & $0.066\,(0.002)$ & $\mathbf{0.062\,(0.002)}$ & $0.092\,(0.003)$ \\
bio       & $0.041\,(0.002)$ & $0.038\,(0.001)$ & $0.040\,(0.001)$ & $0.041\,(0.001)$ & $\mathbf{0.022\,(0.001)}$ & $0.033\,(0.001)$ & $0.033\,(0.001)$ \\
births1   & $0.060\,(0.002)$ & $0.184\,(0.003)$ & $0.092\,(0.002)$ & $0.070\,(0.002)$ & $0.070\,(0.002)$ & $\mathbf{0.070\,(0.001)}$ & $0.091\,(0.002)$ \\
births2   & $0.114\,(0.003)$ & $0.386\,(0.003)$ & $0.065\,(0.002)$ & $0.174\,(0.003)$ & $\mathbf{0.038\,(0.002)}$ & $0.106\,(0.003)$ & $0.057\,(0.002)$ \\
blog\_data& $0.045\,(0.002)$ & $0.199\,(0.003)$ & $0.108\,(0.003)$ & $0.079\,(0.002)$ & $0.058\,(0.002)$ & $\mathbf{0.054\,(0.002)}$ & $0.107\,(0.003)$ \\
calcofi   & $0.050\,(0.002)$ & $\mathbf{0.032\,(0.001)}$ & $0.032\,(0.001)$ & $0.041\,(0.001)$ & $0.033\,(0.001)$ & $0.037\,(0.001)$ & $0.031\,(0.001)$ \\
house     & $0.100\,(0.002)$ & $0.117\,(0.003)$ & $0.098\,(0.002)$ & $0.101\,(0.002)$ & $0.120\,(0.002)$ & $0.110\,(0.002)$ & $\mathbf{0.093\,(0.002)}$ \\
households& $0.146\,(0.003)$ & $0.090\,(0.002)$ & $\mathbf{0.022\,(0.001)}$ & $0.093\,(0.002)$ & $0.093\,(0.002)$ & $0.065\,(0.002)$ & $0.117\,(0.003)$ \\
meps\_19  & $0.050\,(0.002)$ & $0.226\,(0.003)$ & $0.038\,(0.002)$ & $0.064\,(0.002)$ & $0.044\,(0.002)$ & $\mathbf{0.036\,(0.002)}$ & $0.055\,(0.003)$ \\
meps\_20  & $0.045\,(0.002)$ & $0.234\,(0.003)$ & $0.036\,(0.002)$ & $0.066\,(0.002)$ & $0.040\,(0.002)$ & $\mathbf{0.036\,(0.002)}$ & $0.052\,(0.003)$ \\
meps\_21  & $0.052\,(0.002)$ & $0.207\,(0.003)$ & $0.036\,(0.002)$ & $0.062\,(0.002)$ & $0.043\,(0.002)$ & $\mathbf{0.039\,(0.002)}$ & $0.052\,(0.003)$ \\
scm1d     & $0.066\,(0.002)$ & $0.090\,(0.002)$ & $0.088\,(0.002)$ & $0.057\,(0.002)$ & $0.067\,(0.002)$ & $\mathbf{0.043\,(0.002)}$ & $0.099\,(0.003)$ \\
scm20d    & $0.075\,(0.002)$ & $0.086\,(0.002)$ & $0.067\,(0.002)$ & $0.070\,(0.002)$ & $0.074\,(0.002)$ & $\mathbf{0.036\,(0.001)}$ & $0.142\,(0.003)$ \\
scpf      & $0.084\,(0.003)$ & $0.347\,(0.003)$ & $\mathbf{0.068\,(0.003)}$ & $0.123\,(0.003)$ & $0.098\,(0.003)$ & $0.106\,(0.003)$ & $0.113\,(0.003)$ \\
taxi      & $0.081\,(0.002)$ & $\mathbf{0.034\,(0.001)}$ & $0.081\,(0.002)$ & $0.074\,(0.002)$ & $0.065\,(0.002)$ & $0.080\,(0.002)$ & $0.079\,(0.002)$ \\
wage      & $0.137\,(0.003)$ & $0.414\,(0.003)$ & $0.097\,(0.003)$ & $0.130\,(0.003)$ & $\mathbf{0.083\,(0.003)}$ & $0.118\,(0.003)$ & $0.096\,(0.003)$ \\
wq        & $0.246\,(0.003)$ & $0.394\,(0.003)$ & $\mathbf{0.026\,(0.001)}$ & $0.163\,(0.003)$ & $0.113\,(0.003)$ & $0.088\,(0.003)$ & $0.340\,(0.003)$ \\
\bottomrule
\end{tabular}
}
\caption{PCEs when the model is trained with \textbf{PCA+ALL} and evaluated on each simple pre-rank. Each entry shows mean PCE (std) over five runs. Bold indicates the best (lowest) PCE within each dataset row.}
\label{tab:pce_pcaALL_simple_best_3dp}
\end{table}
}

\end{document}